\documentclass[11pt]{article}

\usepackage[final]{acl}

\usepackage{times}
\usepackage{latexsym}

\usepackage[T1]{fontenc}
\usepackage[utf8]{inputenc}

\usepackage{microtype}

\usepackage{inconsolata}

\usepackage{graphicx}

\usepackage{amsmath}
\usepackage{booktabs}   % 专业表格线
\usepackage{multirow}    % 多行合并
\usepackage{siunitx}     % 数字对齐
\usepackage{threeparttable} % 表格注释
\usepackage{algorithm}
\usepackage{algpseudocode}
\usepackage{newfloat}
\usepackage{listings}
\DeclareCaptionStyle{ruled}{labelfont=normalfont,labelsep=colon,strut=off} % DO NOT CHANGE THIS
\floatstyle{ruled}
\newfloat{listing}{tb}{lst}{}
\floatname{listing}{Listing}
\title{Learn to Relax with Large Language Models: Solving Constraint Optimization Problems via Bidirectional Coevolution}

\author{
  Beidan Liu\textsuperscript{1,4},
  Zhengqiu Zhu\textsuperscript{1,4}\textsuperscript{*},
  Chen Gao\textsuperscript{2},
    Tianle Pu\textsuperscript{4},
  Yong Zhao\textsuperscript{1,4},
  Wei Qi\textsuperscript{3},
  Quanjun Yin\textsuperscript{1,4}
  \\
  \small{
  \textsuperscript{1}State Key Laboratory of Digital Intelligent Modeling and Simulation, National University of Defense Technology, Changsha, China}\\
  \small{
  \textsuperscript{2}BNRist, Tsinghua University, Beijing, China
  \textsuperscript{3}Department of Industrial Engineering, Tsinghua University, Beijing, China}\\
  \small{
  \textsuperscript{4}College of Systems Engineering, National University of Defense Technology, Changsha, China}\\
\small{
  \textbf{Email:}
  \{\texttt{liubangdan, zhuzhengqiu12, zhaoyong15, putl22}\}@nudt.edu.cn,}\\
\small{\href{mailto:chgao96@mails.tsinghua.edu.cn}{chgao96@mails.tsinghua.edu.cn},
  \href{mailto:qiw@tsinghua.edu.cn}{qiw@tsinghua.edu.cn},
  \href{mailto:yin_quanjun@163.com}{yin\_quanjun@163.com}
}
}

\begin{document}
\maketitle
\begin{abstract}
Large Language Model (LLM)-based optimization has recently shown promise for autonomous problem solving, yet most approaches still cast LLMs as passive constraint checkers rather than proactive strategy designers, limiting their effectiveness on complex Constraint Optimization Problems (COPs). To address this, we present AutoCO, an end-to-end \textbf{Auto}mated \textbf{C}onstraint \textbf{O}ptimization method that tightly couples operations-research principles of constraint relaxation with LLM reasoning. A core innovation is a unified triple-representation that binds relaxation strategies, algorithmic principles, and executable codes. This design enables the LLM to synthesize, justify, and instantiate relaxation strategies that are both principled and executable. To navigate fragmented solution spaces, AutoCO employs a bidirectional global–local coevolution mechanism, synergistically coupling Monte Carlo Tree Search (MCTS) for global relaxation-trajectory exploration with Evolutionary Algorithms (EAs) for local solution intensification. This continuous exchange of priors and feedback explicitly balances diversification and intensification, thus preventing premature convergence.
Extensive experiments on three challenging COP benchmarks validate AutoCO's consistent effectiveness and superior performance, especially in hard regimes where current methods degrade. 
Results highlight AutoCO as a principled and effective path toward proactive, verifiable LLM-driven optimization. %Codes and source code are available at https://anonymous.4open.science/r/AutoCO1/.

%While recent Large Language Model (LLM)-based optimization methods show promise for autonomous problem-solving, they predominantly function as passive constraint validators rather than proactive strategy designers, failing to handle the sophisticated constraint interactions inherent to constraint optimization problems (COPs). To address these limitations, we introduce the first end-to-end \textbf{Auto}mated \textbf{C}onstraint \textbf{O}ptimization (AutoCO) method, which revolutionizes COP resolution through learning to relax with LLMs. Specifically, we leverage structured LLM reasoning to generate constraint relaxation strategies that evolve dynamically, integrating algorithmic principles and executable code through a unified triple-representation scheme. Additionally, we establish a novel bidirectional (global-local) coevolution mechanism that synergistically combines Evolutionary Algorithms for intensive local refinement with Monte Carlo Tree Search for systematic global strategy space exploration, ensuring an optimal balance between intensification and diversification in fragmented solution spaces. Finally, comprehensive experiments on three challenging COP benchmarks validate AutoCO's consistent effectiveness and superior performance over existing SOTA methods, particularly in scenarios where traditional solvers struggle.
\end{abstract}

%File: formatting-instructions-latex-2026.tex
%release 2026.0

\begin{figure}[t] 
    \centering % 居中对齐
    \includegraphics[width=0.5\textwidth,height=4.5cm, keepaspectratio=false]{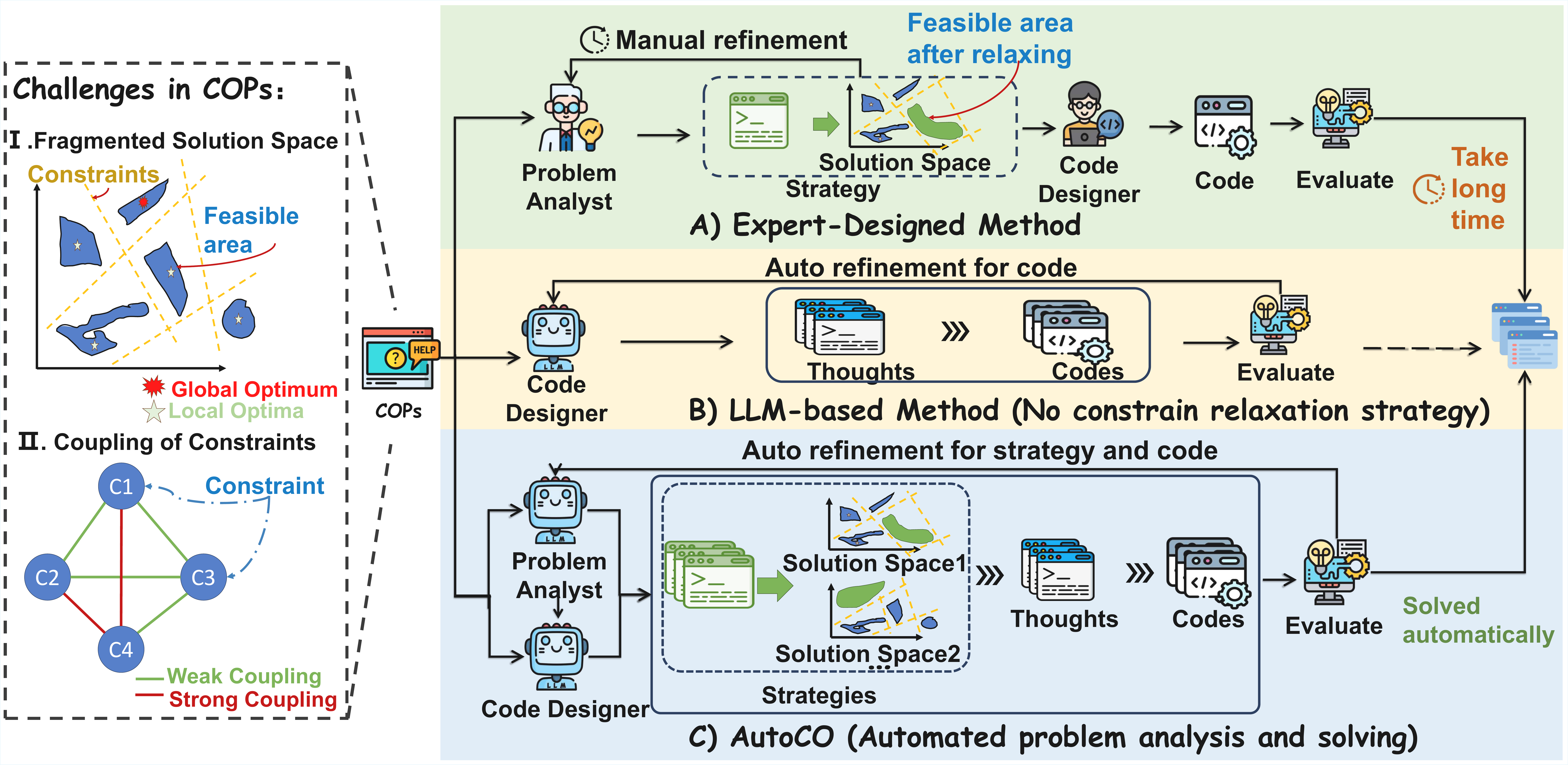} 
    \caption{Comparisons of human/LLM-based solutions for COPs. A) Expert-designed method leverages human analysis to relax constraints for feasible solutions. B) Current LLM-based methods focus on code generation, lacking systematic problem analysis. C) Our AutoCO combines human-inspired relaxation strategies with automation to effectively discover feasible solutions.}
    \label{fig:1} 

\end{figure}
\section{Introduction}

Constraint optimization problems (COPs) are extensively studied in logistics \cite{yaoCooperativeOperationFleet2024}, finance, and industrial planning \cite{mokhtarzadehHybridClusteringMetaheuristic2021}. Most real-world COPs are NP-hard, with complexity arising from hard constraints that create conflicting decisions and global dependencies \cite{1981CombinatorialOA}. These constraints fragment the feasible region, complicating the solution process and underscoring the necessity for efficient and high-quality solutions.

Many methods have been proposed for solving COPs, such as branch-and-bound, heuristics, etc. While these traditional approaches perform well in specific contexts, they often rely heavily on expert experience for effective adjustments to various problems. Recently, Large Language Model (LLM)-based methods have gained attention for their potential to provide end-to-end solutions \cite{liuSystematicSurveyLarge2024a}. These methods leverage pre-trained knowledge and specific evolutionary frameworks to design algorithms tailored for COPs (e.g., Reevo\cite{yeReEvo2024}, EoH\cite{liuLargeLanguageModel2024c}).

However, existing LLM methods primarily rely on feedback from feasible solutions to guide algorithm design toward optimality\cite{romera-paredesMathematicalDiscoveriesProgram2024b}. As hard constraints increase, obtaining feasible solutions becomes more difficult, resulting in ineffective feedback. This can lead to unguided search processes that produce suboptimal designs. Furthermore, these methods prioritize end-to-end generation but lack detailed problem analysis and strategy design, creating a disconnect from human experiences.

To address COPs with hard constraints, researchers often utilize relaxation strategies, starting with a relaxed version of the original problem \cite{constraintadaptation1999}. After identifying partial feasible solutions, they gradually tighten the constraints to find a complete solution. This relaxation approach mitigates the fragmentation of the feasible region caused by hard constraints, facilitating rapid identification of feasible solutions. The concept is widely used in operations research, such as in Large Neighborhood Search, where relaxing constraints enables the systematic exploration of diverse regions in the search space \cite{LNS2019}.

Despite the advantages of this approach, existing relaxation strategies often depend on expert knowledge, making the formulation time-consuming. Integrating the world knowledge, reasoning, and strategy search capabilities of LLMs with these relaxation principles presents a promising opportunity, but introduces significant challenges:
\textit{1) current LLM-based methods lack targeted mechanisms for exploring and relaxing constraints in constraint optimization landscapes, limiting their applicability to large-scale complex scenarios; 2) existing method focus on optimizing code and algorithmic concepts, neglecting the holistic representation of relaxation strategies, which hinders LLM's ability to effectively optimize all three aspects together; and 3) the vast decision space makes it challenging to quickly identify the optimal relaxation strategy.}

To address these challenges, we introduce AutoCO (\textbf{Auto}mated \textbf{C}onstraint \textbf{O}ptimization via LLM-driven bidirectional coevolution), an innovative solution that positions LLMs as proactive strategy designers for COPs (see Figure 1C). AutoCO enables LLMs to explicitly explore and optimize constraint relaxation strategies as integral components in the algorithm design process. The core innovation is a triple-representation scheme that maintains synchronized evolution across three coupled components: constraint relaxation strategies, algorithmic principles, and executable code. This unified representation ensures that each generated solver explicitly employs its corresponding constraint relaxation strategy to solve the optimization problem. To efficiently coordinate this complex co-design process, we develop a bidirectional coevolution mechanism that combines local refinement via Evolutionary Algorithms (EA) with global exploration through Monte Carlo Tree Search (MCTS), enabling effective discovery of performant strategy-concepts-code triples. The contributions of this work are concluded as follows:

%To address the challenges, we introduce AutoCO (\textbf{Auto}mated \textbf{C}onstraint \textbf{O}ptimization via LLM-driven bidirectional coevolution), an innovative solution that enables LLMs to act as proactive strategy designers for COPs (see Figure 1C). Our approach enables LLMs to learn the exploration and formulation of constraint relaxation strategies. One of the innovations is the simultaneous evolution of three coupled components—constraint relaxation strategy, algorithmic principle, and executable code—unified by a novel triple-representation scheme. This enables LLMs to maintain coherence across multiple abstraction levels. To explore the solution space of COPs, we propose a bidirectional coevolution mechanism that combines local refinement via Evolutionary Algorithms (EA) with global exploration through Monte Carlo Tree Search (MCTS). This bidirectional knowledge transfer facilitates strategic adjustments to constraint boundaries, enabling efficient exploration of relaxation strategies.
%Our approach merges constraint relaxation principles from operations research with LLM， encoding expert knowledge into structured principles that allow LLMs to design and optimize relaxation strategies from a problem-oriented perspective

\begin{itemize}
    \item We propose an end-to-end, three-phase solution (i.e., problem analysis, strategy search, and code execution) for solving COPs that empowers LLMs to master the complex art of constraint relaxation.
    %We propose an LLM-driven hybrid heuristic search framework based on constraint relaxation principles, thereby expanding the capability of LLMs to effectively tackle constraint optimization challenges.
    \item We present a triple-representation scheme that maintains synchronized evolution of constraint strategies, algorithmic concepts, and executable code across abstraction levels. This unified encoding bridges the structural modeling gap in conventional approaches.
    \item We develop a bidirectional coevolution mechanism to address the vast strategy decision space. It delegates the fine-grained optimization of strategy-concepts-code triples to the local EA layer, while the global MCTS layer explores the global strategy space to identify promising directions.
    
    \item We validate AutoCO's effectiveness on three challenging COP benchmarks, showcasing its ability to tackle complex constraints, especially where Gurobi struggles, achieving a 24.7\% average optimality gap reduction compared to the SOTA LLM-based methods.
\end{itemize}

\begin{figure}[t] 
    \centering % 居中对齐
    \includegraphics[width=0.5\textwidth,height=4.5cm, keepaspectratio=false]{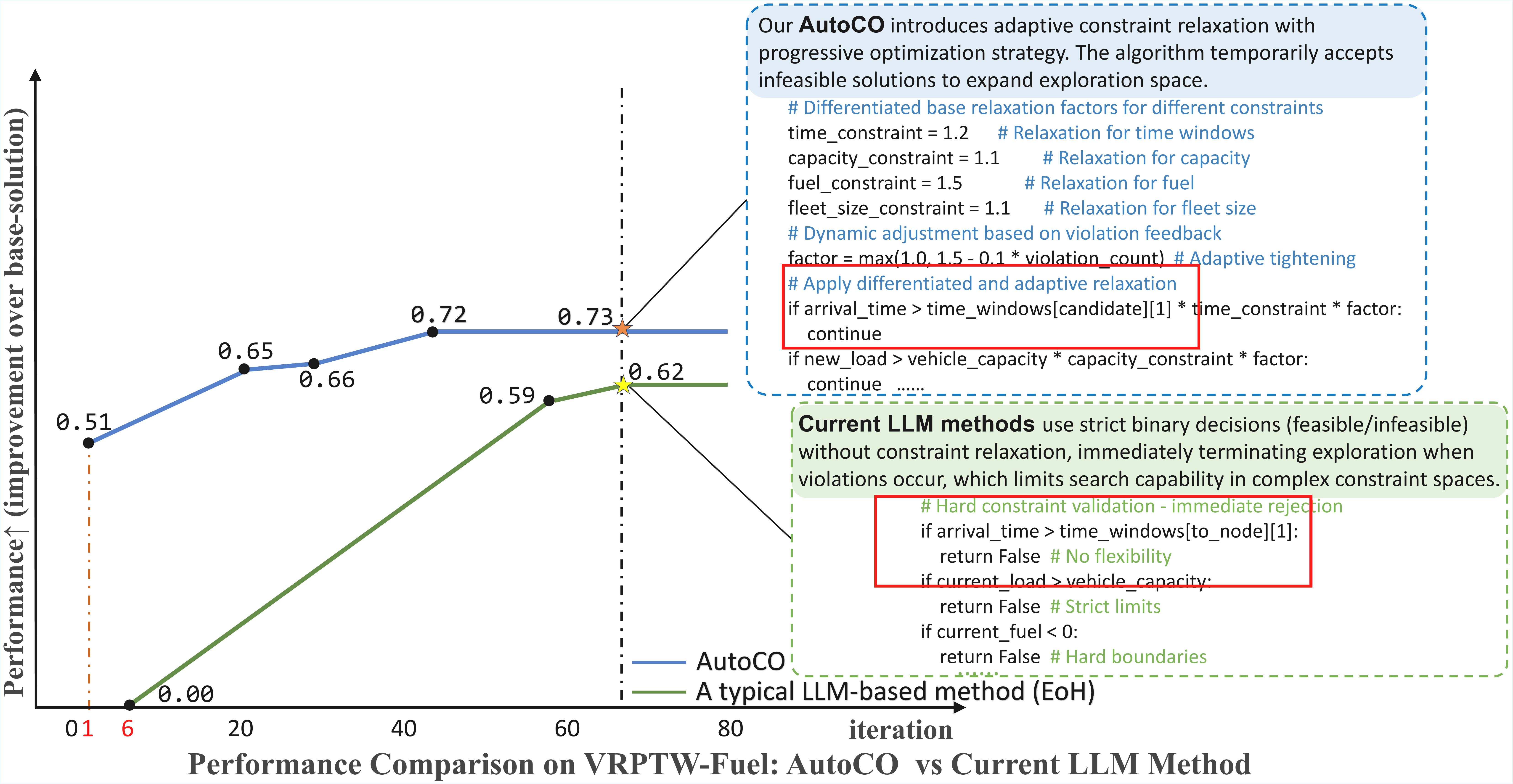} 
    \caption{AutoCO (blue) vs. Current LLM method (green) performance on VRPTW-fuel problems over 100 iterations. The autonomous constraint relaxation temporarily expands feasible regions, enhancing optimization feedback.}
    \label{fig:2} 

\end{figure}
%Performance comparison and algorithmic design analysis between AutoCO (blue) and current end-to-end LLM-based methods (green) on VRPTW-fuel problems. The line chart displays fitness value ratios relative to baseline across 100 iterations. Our AutoCO enables LLMs to autonomously design constraint relaxation strategies, temporarily expanding the feasible region and increasing the probability of obtaining effective feedback.

\section{Related Works}

\subsection{Traditional Methods for Constraint Relaxation}
Various constraint relaxation methods have been developed to simplify complex operations research problems. Linear relaxation transforms integer programs into linear problems by relaxing integrality constraints \cite{linerrelaxation2025}, while Lagrangian relaxation dualizes hard constraints into the objective via multipliers \cite{Lagrangianrelaxation1981}. Recent works embed relaxation concepts directly into algorithm design: Kiessling temporarily relax constraint boundaries to boost performance \cite{Kiessling2024AnAF}, whereas Cao gradually relax constraints to improve search efficiency \cite{sequentialconstraintrelaxation}. \textit{However, these methods rely on expert knowledge, limiting their adaptability in different scenarios \cite{Tseng1987}.}
%To simplify complex operations research problems, various methods for constraint relaxation have been developed. One notable approach is linear relaxation, transforming integer optimization problems into linear programming problems by relaxing integer constraints to continuous variables \cite{linerrelaxation2025}. Similarly, Lagrangian relaxation incorporates challenging constraints into the objective function using Lagrange multipliers \cite{Lagrangianrelaxation1981}. Additionally, recent works have integrated relaxation concepts into algorithm design. For instance, Kiessling’s approach temporarily relaxes constraint boundaries to enhance algorithm performance \cite{Kiessling2024AnAF}. In contrast, Cao's method focuses on gradually relaxing constraints, which improves search efficiency for optimal solutions \cite{sequentialconstraintrelaxation}. \textit{However, these methods require expertise, limiting adaptability in changing scenarios \cite{Tseng1987}.}

\subsection{LLMs for Optimization Problems}
LLMs are applied to optimization in diverse roles \cite{liuSystematicSurveyLarge2024a}, serving as predictors for solution quality \cite{surrogatemodelsHAO2024,NEURIPS2023_d862f7f5} or as feature extractors for domain knowledge \cite{Wuenhancedalgorithm2024}. While demonstrating potential, these passive roles restrict reasoning and optimization. Research thus explores LLMs as active optimizers for solution generation \cite{yangLargeLanguageModels2024} and even as algorithm designers \cite{romera-paredesMathematicalDiscoveriesProgram2024b,liuEvolutionHeuristicsEfficient2024a,yeReEvo2024}, often integrated with evolutionary computation. \textit{Nevertheless, existing LLM methods lack effective problem analysis for COPs, hampering their performance under hard constraints.}

%The current research explores the diverse roles of LLMs in optimization problems \cite{liuSystematicSurveyLarge2024a}. LLMs serve as predictors in regression and classification to forecast solution quality \cite{surrogatemodelsHAO2024,NEURIPS2023_d862f7f5}, and as feature extractors to uncover domain knowledge \cite{Wuenhancedalgorithm2024}. These applications demonstrate LLMs' potential, but their passive roles limit reasoning and optimization capabilities. Researchers are exploring LLMs as active optimizers, enabling solution generation \cite{yangLargeLanguageModels2024}. An emerging approach employs LLMs as algorithm designers \cite{romera-paredesMathematicalDiscoveriesProgram2024b,liuEvolutionHeuristicsEfficient2024a,yeReEvo2024}, integrating with evolutionary computation to address interpretability challenges. \textit{However, existing LLM methods lack effective handling of problem perspectives when addressing COPs, limiting their optimization capabilities under multiple hard constraints.}

\subsection{Synergistic Mechanisms of Evolutionary Algorithms and Monte Carlo Tree Search}

EA and MCTS are widely used for optimization: EA explores globally via population evolution \cite{EibenIntroduction2015}, while MCTS structures search through trees \cite{BrowneSurveyofMonte2012}. Their combination is studied in multi-objective optimization \cite{HongImprovingPerformance2024}, game action evolution \cite{BaierEvolutionaryMCTS2018}, and space decomposition \cite{xiaLearningSearchPromising2022}. \textit{However, these hybrids remain largely domain-specific, focusing on narrow classes like games or continuous spaces \cite{baiWindFarmLayout2022}, limiting universal applicability.}

%EA and MCTS are widely adopted for optimization problems. EAs perform global exploration via population evolution \cite{EibenIntroduction2015}, while MCTS organizes searches through tree structures \cite{BrowneSurveyofMonte2012}. Recent research has explored their synergistic potential, including multi-objective optimization \cite{HongImprovingPerformance2024}, game scenario action sequence evolution \cite{BaierEvolutionaryMCTS2018}, and space decomposition in constrained search problems \cite{xiaLearningSearchPromising2022}. \textit{Critically, these hybrid approaches remain predominantly domain-specific, primarily focusing on narrow problem classes like multi-action games, continuous optimization spaces, or specific optimization problems \cite{baiWindFarmLayout2022,xiaLearningSearchPromising2022}, thereby constraining their universal applicability and generalization potential.}

\section{Preliminaries}
\subsection{Problem Definition of COPs}
\label{subsec:problem-and-relaxation}

COPs can be formally modeled as follows:

\begin{equation}
\begin{aligned}
(\mathrm{P}):\min_{\boldsymbol{x}} \quad & f(\boldsymbol{x}) \label{eq:obj} \\
\text{s.t.} \quad & g_k(\boldsymbol{x}) \leq 0, \quad k = 1, \dots, m \\ 
& h_j(\boldsymbol{x}) = 0, \quad j = 1, \dots, p \\
& \boldsymbol{x} \in \mathcal{X},
\end{aligned}
\end{equation}
where $f(\boldsymbol{x})$ is the objective function, $g_k(\boldsymbol{x})$ are inequality constraints, $h_j(\boldsymbol{x})$ are equality constraints, and $\boldsymbol{x}$ is the vector of decision variables. The set $\mathcal{X}$ defines the feasible region for $\boldsymbol{x}$, encompassing both continuous and discrete representations.

\subsection{Constraint Relaxation Strategy}
\label{subsubsec:relaxation-strategy}
%To effectively address the challenges associated with COPs, we employ a constraint relaxation strategy informed by prior research \cite{grossmannReviewNonlinearMixedinteger,floudasNonlinearMixedintegerOptimization1995}. This strategy expands the feasible region by temporarily relaxing constraints, thereby enhancing the efficiency of the solution search. The relaxation strategy can be expressed as:
%This strategy aims to approximate the feasible region and enable the exploration of solutions more efficiently. The relaxation strategy can be expressed as:

To enable explicit exploration and optimization of constraint relaxation strategies, we formalize them as a structured search space based on prior research \cite{grossmannReviewNonlinearMixedinteger,floudasNonlinearMixedintegerOptimization1995}. A constraint relaxation strategy $\sigma$ is defined as a set of pairs, each specifying a constraint and its corresponding relaxation factor:
\begin{equation}
\sigma = \{ (g_1, \delta_1), \dots, (g_m, \delta_m) \}, \label{eq:relaxation}
\end{equation}
where $g_k$ is the $k$-th constraint and $\delta_k \in [\alpha_k, \beta_k]$ its relaxation factor, with $\alpha_k$, $\beta_k$ defining the lower and upper relaxation bounds. 

The strategy space $\mathcal{T}$ is defined as the set of all possible relaxation strategies:
\begin{equation}
\mathcal{T} = \{ \sigma_1, \sigma_2, \dots, \sigma_N \}. \label{eq:strategy-space}
\end{equation}

%The initial optimization problem (\ref{eq:obj}) can subsequently be reformulated into a relaxed problem:
Applying a relaxation strategy $\sigma$ to the original problem yields a relaxed problem:

\begin{equation}
\begin{aligned}
(\mathrm{P}_\sigma): \min_{\boldsymbol{x}}  \quad& f(\boldsymbol{x}) \label{eq:relaxation-p}\\
\text{s.t.} \quad & g_k(\boldsymbol{x}) \leq \delta_k, \quad k = 1, \dots, m  \\
             & h_j(\boldsymbol{x}) = 0,  \quad j = 1, \dots, p \\
& \boldsymbol{x} \in \mathcal{X}.
\end{aligned}
\end{equation}

This transformation broadens the feasible region ($\mathcal{F}_{\mathrm{P}_\sigma} \supseteq \mathcal{F}_{\mathrm{P}}$) when $\delta_k \geq 0$, facilitating initial solution discovery. Crucially, different relaxation strategies $\sigma \in \mathcal{T}$ produce distinct expanded regions and search trajectories, potentially influencing initial solution efficiency and final solution quality in $\mathcal{F}_{\mathrm{P}}$.

\section{Methodology}
\subsection{Overview}
\label{subsec:framework}
%AutoCO leverages constraint relaxation strategies designed and optimized by LLMs to emulate human-like capabilities in addressing relaxation optimization problems, thereby tackling the challenges posed by an increasing number of hard constraints in combinatorial optimization problems. As illustrated in Figure \ref{fig:framework}, AutoCO consists of three stages: problem analysis and initial strategy design, optimal strategy search, and code execution.
%\textbf{In the first stage}, our approach parses problem descriptions to generate an initial set of constraint relaxation strategies. Each strategy aims to intelligently adjust the constraint boundaries of the COPs to temporarily increase the probability of obtaining feasible solution feedback, thereby enhancing the exploration of viable solutions. \textbf{The optimal strategy search stage} employs a dual-layer cooperative optimization mechanism: the inner layer utilizes EA for fine-grained optimization, while the outer layer leverages MCTS to comprehensively explore the global strategy space, ensuring thorough search coverage. \textbf{The code execution stage evaluates} the generated algorithms on problem instances to provide feedback for further optimization (Detailed configurations are in Appendix A).

\begin{figure*}[t]
    \centering
    \includegraphics[width=\linewidth,height=7cm, keepaspectratio=false]{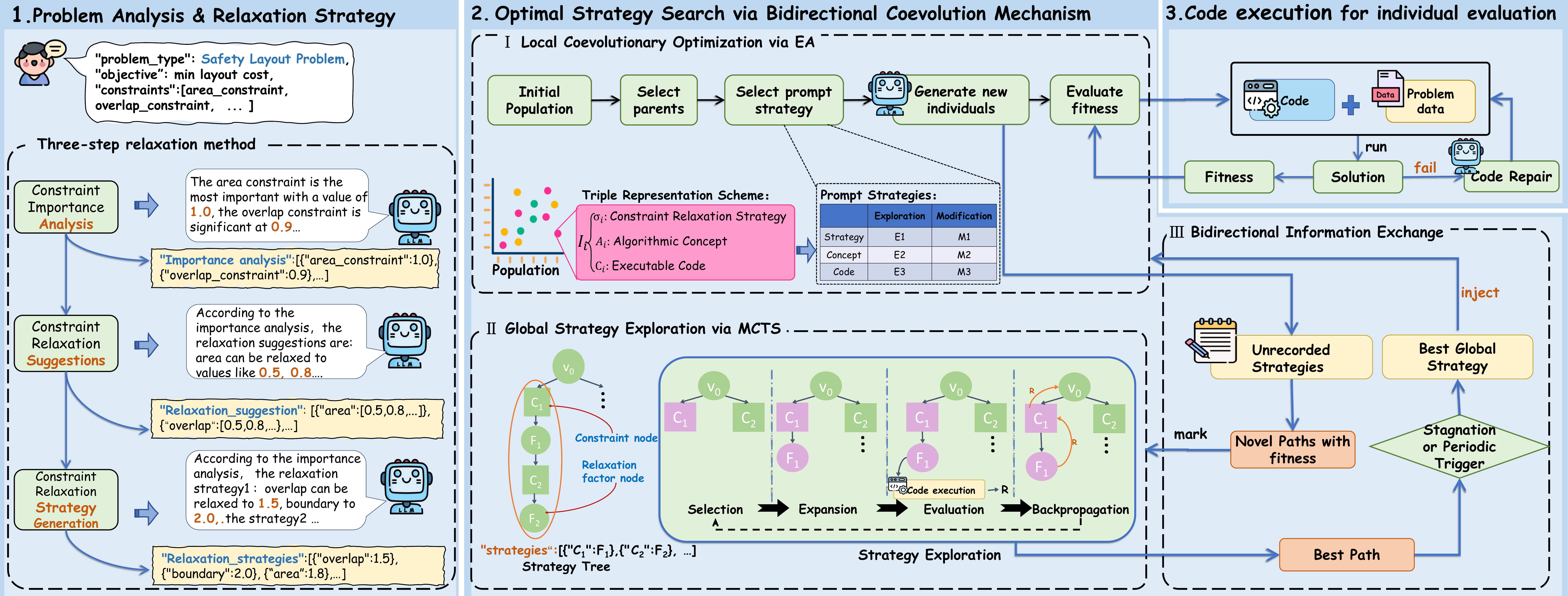}
    \caption{Architecture of AutoCO. Initially, we use LLMs to parse user-input problems and generate initial constraint relaxation strategies. Next, the bidirectional coevolution mechanism combining local EA and global MCTS explores and optimizes strategies and codes. Finally, we evaluate the generated algorithms on problem instances and provide individual fitness feedback.}
    \label{fig:framework}
\end{figure*}

AutoCO leverages constraint relaxation strategies designed and optimized by LLMs to exhibit human-like capabilities in addressing constraint optimization problems, thereby tackling the challenges posed by increasing hard constraints in constraint optimization. As illustrated in Figure \ref{fig:framework}, AutoCO comprises three stages: problem analysis and initial strategy design, optimal strategy search, and code execution.

\textbf{In the first stage}, our approach parses problem descriptions to generate an initial set of constraint relaxation strategies. Each strategy modifies the constraints of the COPs to temporarily enhance the probability of searching feasible solutions, thereby fostering a more extensive exploration of potential solutions. \textbf{The optimal strategy search stage} employs a dual-layer cooperative optimization mechanism: the local layer utilizes EA for fine-grained optimization, while the global layer leverages MCTS to explore the strategy space thoroughly, ensuring comprehensive search coverage. \textbf{The code execution stage} evaluates generated algorithms on problem instances to provide feedback for further optimization (Details about this stage are provided in Appendix A).

\subsection{LLM-Driven Constraint Relaxation Strategy Generation}

\label{subsec:strategy_generation}
 To address the limitations of expert reliance and to incorporate the insight of constraint relaxation, we propose a three-step relaxation method that transforms expert-dependent constraint handling into an LLM-driven automated approach, by combining domain expertise with LLM reasoning capabilities.

\paragraph{Constraint Importance Analysis:} Constraint relaxation strategies grow exponentially with the increasing number of problem constraints.  We utilize LLMs to parse problem text to identify all constraints $\mathcal{G} = \{g_1, g_2, \dots, g_m\}$ and strategically assign importance weights $w_i \in [0, 1]$ to each constraint $g_i$ (with $w_i=1$ for critical constraints and $w_i=0$ for negligible ones), enabling a shift from exhaustive enumeration to intelligent, context-aware constraint analysis.

\paragraph{Constraint Relaxation Range Suggestion:} Based on constraint analysis, we determine adaptive relaxation factor ranges $[\alpha_i, \beta_i]$ ($\alpha_i \leq 1 \leq \beta_i$) for each constraint $g_i$ using LLMs. This range defines permissible relaxation factors that can be applied to each constraint: $1$ corresponds to the original constraint boundary, $\alpha_i < 1$ permits constraint tightening, and $\beta_i > 1$ allows controlled constraint violation.

\paragraph{Constraint Relaxation Strategy Generation:} The final step synthesizes insights from the weights $W = \{w_1, \dots, w_m\}$ and ranges $R = \{[\alpha_1,\beta_1], \dots, [\alpha_m,\beta_m]\}$, the LLM generates a initial strategy set $\Sigma = \{\sigma_1,\dots, \sigma_k\}$, where each strategy $\sigma_j = (\delta_{j1}, \dots, \delta_{jm})$ is an $m$-dimensional vector representing relaxation coefficients. Crucially, this sampling process adheres to constraint criticality structures, ensuring that the generated strategies are derived from the systematic analysis of constraint importance and relaxation potential, rather than produced randomly.

\subsection{Triple Representation Scheme}

%Existing LLM-based optimization methods \cite{liuEvolutionHeuristicsEfficient2024a} using dual representations (algorithm and code) struggle to generate feasible COP solutions under challenging constraints, leading to high computational costs and inefficiencies.

Existing LLM-based optimization methods \cite{liuEvolutionHeuristicsEfficient2024a} use single or dual representations (algorithm and code) but neglect to generate and optimize constraint relaxation strategies. As hard constraints increase, these methods fail to provide timely feedback, resulting in high computational costs and inefficiencies.

To address this limitation, we design a novel triple individual representation by incorporating constraint relaxation strategies. Our comprehensive triple representation scheme $\mathcal{I}_j = \langle \sigma_j, A_j, C_j \rangle$ explicitly integrates constraint relaxation strategy $\sigma_j$, algorithmic concept $A_j$, and executable code $C_j$. This novel representation provides a systematic mechanism for strategy-code coevolution, transcending traditional boundaries between constraint handling, design, and implementation. By facilitating synchronized evolution across different abstraction levels, our approach enables rapid optimization and verification of relaxation strategies for COPs. An example scheme can be found in Appendix A.

\subsection{Optimal Strategy Search via Bidirectional Coevolution Mechanism}
\label{subsec:strategy_search}
To address complexities from hard constraints, particularly the rapid identification of effective strategies amid fragmented solution spaces and the extensive decision space of relaxation strategies, we devise a bidirectional coevolution mechanism combining a local EA with a global MCTS:
%To mitigate the risk of local optima and balance local optimization with global strategy exploration, we devise a bidirectional coevolution mechanism combining a local EA with a global MCTS:
\begin{equation}
\mathcal{M}_\text{exchange} = \text{Bidirec\_Update}(\mathcal{T}_\text{MCTS}, \mathcal{P}_\text{EA}),
\end{equation}
where $\mathcal{M}_\text{exchange}$ represents the bidirectional exchange mechanism, $\mathcal{T}_\text{MCTS}$ is the MCTS search tree, and $\mathcal{P}_\text{EA}$ is the EA population.

The mechanism leverages MCTS's global perspective to help EA escape local optima, enabling effective navigation and optimization across diverse solution regions. Furthermore, it enhances resource efficiency by integrating EA's exploration results into MCTS. It comprises three key components: (1) local coevolutionary optimization, (2) global strategy exploration, and (3) bidirectional information exchange. %, as detailed in Appendix A.

\paragraph{Local Coevolutionary Optimization via EA}
\label{subsubsec:local_ea}
The EA performs local evolutionary refinement to explore promising solutions. Parent selection uses tournament-based competition to balance selection pressure and diversity:
%\begin{equation}
%\mathcal{P}_{EA}^{t+1} = \text{Evolution}(\mathcal{P}_{EA}^t; \lambda),
%\end{equation}
%where $\mathcal{P}_{EA}^{t+1}$ represents the next-generation population and $\lambda$ captures adaptive evolutionary strategies.
\begin{equation}
\mathcal{P}_{EA}^{t+1} = \text{Evolution}(\mathcal{P}_{EA}^t),
\end{equation}
where $\mathcal{P}_{EA}^{t+1}$ represents the next-generation population.
A key innovation is LLM-guided coordinated evolution, modifying constraint relaxation strategies, algorithmic concepts, and executable code while preserving coherence. Specialized prompts generate offspring that maintain logical consistency across the three representation levels. Details of these prompts are in Appendix B. 

\paragraph{Global Strategy Exploration via MCTS}
\label{subsubsec:global_mcts}
The MCTS component explores the constraint relaxation strategy space $\mathcal{T}$ for global guidance. To improve the recording and searching of these strategies, we modify the structure to alternate between constraint nodes $D$ and relaxation factor nodes $R$, where each complete path defines a strategy $\sigma$.

We introduce corresponding UCT calculation formulas for these node types to balance exploration and exploitation during the selection phase:

%\begin{equation}
%\text{UCT}(n) =
%\begin{cases}
%\frac{Q(n)}{N(n)} + k \sqrt{\frac{\ln N(F)}{N(n)}},  \small{n = D_i} \\
%\frac{Q(n)}{N(n)} + k \sqrt{\frac{\ln N(D_i)}{N(n)}} \cdot w_i,  \small{n = R_{ij}}
%\end{cases}
%\end{equation}

\begin{equation}
\text{UCT}(n)\!\! =\!\!
\left\{
\begin{array}{@{}l@{\hspace{-0.1em}}l@{}}
\frac{Q(n)}{N(n)} + k \sqrt{\frac{\ln N(F)}{N(n)}}, & n = D_i \\
\frac{Q(n)}{N(n)} + k \sqrt{\frac{\ln N(D_i)}{N(n)}} \, w_i, & n = R_{ij}
\end{array}
\right.
\end{equation}
where $Q$ is the cumulative rewards for nodes; $N$ is visit counts; $F$ is the parent node; $k$ is the exploration constant balancing exploration and exploitation; $D_i$ is the $i$-th constraint node; and $R_{ij}$ is the $j$-th relaxation factor of the $i$-th constraint.

During the evaluation phase, we use LLM to generate executable code $C_{\text{strategy}}$ for new strategies, with the fitness from code execution serving as the reward $R$ for backpropagation to update node values. Upon search completion, MCTS outputs the optimal strategy $\sigma^*$ to guide the local EA.

\paragraph{Bidirectional Information Exchange}
\label{subsubsec:bidirectional_flow}

The bidirectional exchange mechanism integrates EA and MCTS through strategic information transfer, enhancing computational efficiency and aiding in overcoming local optima.

\begin{equation}
\mathcal{I}_{\text{exchange}} = \left\{\begin{array}{l}
\text{S}_{\text{MCTS}} \to \mathcal{P}_{\text{EA}} \\
\text{I}_{\text{EA}} \to \mathcal{T}_{\text{MCTS}}
\end{array}\right.,
\end{equation}
where $\text{S}_{\text{MCTS}}$ represents promising strategies from MCTS, $\text{I}_{\text{EA}}$ denotes feasible solutions from EA.

(1) EA→MCTS: Continuous feedback allows each feasible solution produced by EA to update MCTS tree statistics with evaluated constraint-relaxation strategies. This integration diminishes the computational burden on MCTS by utilizing pre-assessed strategies, thus avoiding redundant exploration of known areas.

(2) MCTS→EA: The injection triggers under two conditions: upon detecting EA stagnation (via fitness-based monitoring) or periodically, to sustain global search momentum. The innovation lies in conveying MCTS-derived promising strategies, steering EA toward unexplored areas with high potential. MCTS offers a global perspective by pinpointing strategy combinations that EA's local search may miss, fostering escape from local optima through informed population diversification.

\section{Experiment}
\subsection{Experimental Settings}
\paragraph{Problems and datasets}
To validate AutoCO’s generalizability, we select three COPs that demonstrate different constraint characteristics: Vehicle Routing Problem with Time Windows (VRPTW), VRPTW with fuel (VRPTW-Fuel), and Safety Facility Layout (SFL).

Dataset: VRPTW and VRPTW-fuel (Solomon benchmark: S(25)/M(50)/L(100) customers \cite{solomon1987}), SFL (MINLPLib: 4/8/5(Dual) with safety regions \cite{bonamiAlgorithmicFrameworkConvex2008}).

\begin{table}[h]
\scriptsize
\centering
\caption{Key Challenging Constraints in COPs}
\label{tab:challenging_constraints}
\begin{tabular}{c p{2.5cm} p{2.5cm}}
\toprule
Problem & Constraint & Description \\
\midrule
\multirow{2}{*}{SFL} 
& Non-overlap: $(x_i + w_i \le x_j) \lor (x_j + w_j \le x_i) \lor \cdots$ & Nonlinear geometric exclusion \\
& Safety inclusion: $(x_i, y_i, w_i, h_i) \subseteq \text{circle } a$ & Coupling of continuous placement with discrete regions \\
\addlinespace
\multirow{1}{*}{VRPTW} 
& Time windows: $t_j \ge t_i + s_i + d_{ij} - M(1 - y_{ij})$ & Early decisions constrain later feasibility \\
\addlinespace
\multirow{1}{*}{VRPTW-Fuel } 

& Fuel: $F_r = \sum d_{ij}(c_d + c_\ell L_i) \le F_{\text{cap}}$ & Resource accumulation coupled with load and distance \\

\bottomrule
\end{tabular}
\end{table}

\begin{table*}[t]
\centering
\caption{Performance comparisons with the SOTA baselines (reflected by optimality gap with SOTA,↓).}
\label{tab:gap_comparison}
\scriptsize
\setlength{\tabcolsep}{2.5pt}
\renewcommand{\arraystretch}{1.2}

\begin{tabular}{@{}ll
                *{3}{>{\centering\arraybackslash}p{1.2cm}}
                *{3}{>{\centering\arraybackslash}p{1.2cm}}
                *{3}{>{\centering\arraybackslash}p{1.2cm}}@{}}
\toprule
\multirow[c]{2}{*}{\textbf{Type}} & \multirow[c]{2}{*}{\textbf{Method}} & 
\multicolumn{3}{c}{\textbf{VRPTW}} & 
\multicolumn{3}{c}{\textbf{VRPTW-fuel}} & 
\multicolumn{3}{c}{\textbf{SFL}} \\
\cmidrule(lr){3-5} \cmidrule(lr){6-8} \cmidrule(l){9-11}
 & & \textbf{S} & \textbf{M} & \textbf{L}& \textbf{S} & \textbf{M} & \textbf{L}& \textbf{4} & \textbf{8} & \textbf{5(Dual)} \\
\midrule
Exact Solver & Gurobi & \textbf{0.00} & \textbf{0.00} & \textbf{0.00} &\textbf{0.00} & - & -& \textbf{0.00} & \textbf{0.00} & \textbf{0.00} \\
\midrule
Reinforcement Learning & DeepACO &$\underline{0.29}({0.04})$ &$0.46({0.01})$ & $0.50({0.09})$&$1.12({0.08})$ & $0.97({0.09})$ & $0.38({0.14})$ &$0.04({0.02})$& $0.19({0.03})$ & $0.18({0.02})$\\
\midrule
\multirow[c]{5}{*}{Metaheuristic} 
& MA &$0.45({0.11})$&$0.49({0.10})$ & $0.50({0.50})$ & $0.46({0.08})$&$0.54({0.10})$ & -&$\underline{0.02}({0.03})$ & $0.15({0.03})$& $0.16({0.02})$ \\
& DE & $0.37({0.15})$&$0.40({0.12})$&$0.57({0.10})$&$0.46({0.09})$&$0.89({0.10})$&$0.60({0.06})$&$0.03({0.03})$&$0.14({0.02})$& $0.14({0.02})$\\
& SA &$0.37({0.14})$&$0.36({0.13})$&$0.47({0.14})$&$0.74({0.09})$&$0.84({0.07})$ & $0.72({0.60})$&$0.05({0.03})$& $\underline{0.08}({0.03})$& $\underline{0.08}({0.03})$\\
& GA &$0.43({0.13})$&$0.44({0.06})$&$0.55({0.14})$&$0.91({0.04})$&$0.81({0.09})$ & $0.33({0.15})$&$0.05({0.03})$& $0.16({0.03})$& $0.16({0.03})$\\
& PSO&$0.42({0.14})$&$0.44({0.08})$&$0.47({0.03})$& $0.48({0.04})$ &$0.82({0.11})$ &$\underline{0.31}({0.07})$&$0.09({0.02})$& $0.11({0.02})$ & $0.11({0.03})$\\
\midrule
\multirow{3}{*}{LLM-based} 
& FunSearch& $0.50({0.22})$&$0.80({0.36})$ &$0.70({0.26})$ & $1.30({0.85})$ &$0.45({0.09})$ & - &$0.28({0.06})$& $0.27({0.03})$& $0.28({0.05})$\\
& EoH      & $0.36({0.16})$&$0.63({0.34})$ &$0.52({0.30})$ &$0.81({0.46})$&$\underline{0.20}({0.22})$ & -&$0.18{0.05}$& $0.24({0.03})$ &$0.24({0.07})$\\
& ReEvo    & $0.36({0.18})$&$\underline{0.33}({0.09})$ &$0.50({0.27})$&$0.36({0.25})$ &$\underline{0.20}({0.26})$ & -&$0.13({0.03})$& $0.25({0.03})$ &$0.26({0.04})$\\
\midrule
\textbf{Ours }
& \textbf{AutoCO}&$0.53({0.17})$&$0.45({0.12})$&$\underline{0.42}({0.15})$&$\underline{0.31}({0.07})$ &$\textbf{0.00}({0.08})$ & $\textbf{0.00}({0.02})$&$\underline{0.02}({0.04})$& $0.17({0.03})$ &$0.15({0.02})$\\
\bottomrule
\end{tabular}

\vspace{0.5em}
\begin{minipage}{\textwidth}
\footnotesize
Note: The performance on three problems is reported as mean(std). \textbf{Bold} values represent the optimal solution in each column, while \underline{underlined} values indicate the second-best solution.
'-' denotes no feasible solution found within the limited time.
A lower optimality gap indicates superior performance, reflecting the method's ability to generate high-quality solutions closer to the optimal benchmark.
\end{minipage}
\end{table*}

\paragraph{Baseline Methods} Baselines include exact solver (Gurobi), reinforcement learning (DeepACO \cite{ye2023deepaco,berto2025rl4co}), metaheuristics (SA \cite{SA1983}, GA \cite{GA1994}, PSO \cite{PSO1995}, MA \cite{MA1989}, DE \cite{DE2019}), and LLM methods (FunSearch \cite{romera-paredesMathematicalDiscoveriesProgram2024b}, EoH \cite{liuEvolutionHeuristicsEfficient2024a}, ReEvo \cite{yeReEvo2024}) . 

\paragraph{Evaluation Metrics} Optimality gap $\gamma = \frac{|f_{\mathrm{best}} - f_{\mathrm{opt}}|}{|f_{\mathrm{opt}}|}$, runtime ($T_{e2e}$), feasible solution time ($T_{tff}$), and stagnation time ($T_{stag}$). 
\paragraph{Implementation Settings} DeepSeek-R1, population 45, 2-hour limit, 100 runs on Intel i5-13400F/RTX 4060 Ti.

Details of problem models, datasets, baseline methods, and implementation settings are provided in Appendix C.

% ===== PERFORMANCE COMPARISON =====
\subsection{Comprehensive Performance Comparison}
\label{subsec:overall}
To validate AutoCO's generalizability and effectiveness in solving COPs, we adopt the optimality gap as the primary evaluation metric, comparing AutoCO and baseline methods against state-of-the-art (SOTA) solutions as shown in Table \ref{tab:gap_comparison}.

\begin{itemize}
\item \textbf{Comparison with Traditional Methods}:
Our method demonstrates competitive performance across various optimization domains. Specifically, in the VRPTW domain, AutoCO achieves optimality gaps of 0.53 (S), 0.45 (M), and 0.42 (L) on average. While slightly behind traditional methods in the first two instances, AutoCO shows improvement in larger problem scales, even surpassing methods like MA (0.50) and SA (0.47). %This represents a  improvement over traditional metaheuristic approaches such as SA, which exhibits gaps of 0.37 (S), 0.36 (M), and 0.47 (L) respectively for different scales.
Regarding VRPTW-fuel problem, which introduces additional constraints related to fuel consumption, AutoCO maintains optimality gaps of 0.31 (S), 0.00 (M), and 0.00 (L) while other methods such as DeepACO struggles, showing gaps of 1.12 (S) and 0.97 (M). Notably, Gurobi fails to find feasible solutions for larger instances within the designated time, highlighting AutoCO’s robustness in handling increased complexity and its capability to generate solutions without extensive manual fine-tuning.

\item\textbf{Comparison with LLM-based Methods}:
Compared to existing LLM-based methods, AutoCO demonstrates substantial advantages. In the VRPTW-fuel domain, AutoCO achieves optimality gaps of 0.31 (S) and 0.00 (M), significantly outperforming ReEvo’s gaps of 0.36 (S) and 0.20 (M) respectively. Other LLM methods like FunSearch show a dramatic decline in performance under increased constraints, with a gap of 1.30 (S). Similar tendencies can be observed in other two constrained optimization problems. Our results indicate an average reduction in the optimality gap across problem sets, confirming AutoCO's consistent generation of high-quality solutions as problem complexity escalates, thereby validating the effectiveness of our approach in solving challenging COPs.

\end{itemize}
Our comprehensive analysis reveals that AutoCO significantly advances LLM capabilities in solving COPs. \textit{This performance demonstrates AutoCO's potential to autonomously design strategies, reducing manual engineering and advancing end-to-end automation in COPs. More importantly, AutoCO provides effective initial feasible solutions, assisting traditional methods or exact solvers to search more efficiently for optimal solutions, avoiding waste and stagnation in infeasible regions.}
%Our comprehensive analysis reveals that AutoCO significantly advances the capability of LLMs in solving COPs. \textit{This performance demonstrates AutoCO's potential to autonomously design strategies, reducing manual engineering and advancing end-to-end automated problem-solving in COPs.}

\begin{figure}[htb]
\centering
\includegraphics[width=\columnwidth]{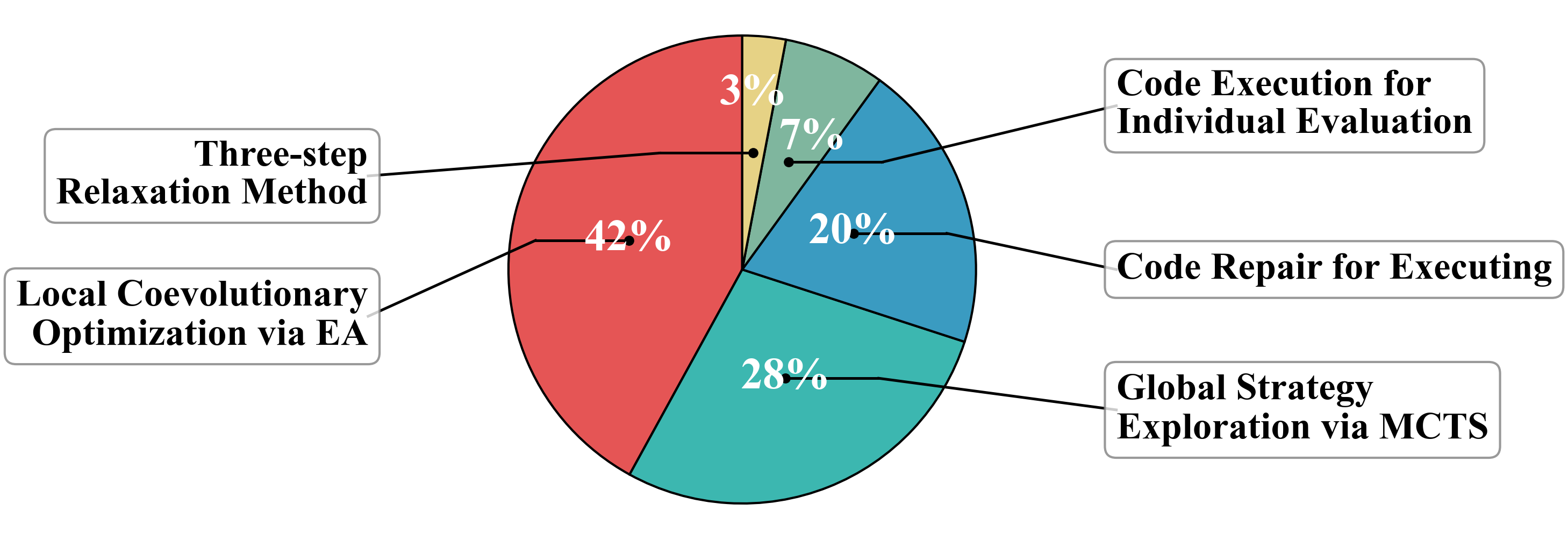}
\caption{Time distribution of AutoCO components.}
\label{fig:time}
\end{figure}

\subsection{Time Efficiency Analysis}
\label{subsec:time_efficiency}

We assess AutoCO's computational performance on VRPTW through three metrics: end-to-end runtime ($T_{e2e}$), time to first feasible solution ($T_{tff}$), and performance stagnation time ($T_{stag}$). As shown in Table \ref{tab:time_mdd}, our temporal analysis reveals nuances in the computational characteristics of our approach compared to three baseline methods. Figure \ref{fig:time} provides a breakdown of AutoCO's internal time distribution across different stages.
\begin{itemize}
    \item \textbf{Efficient Solution Generation and Stagnation Mitigation}: AutoCO demonstrates impressive efficiency in generating initial feasible solutions. As indicated in Table \ref{tab:time_mdd}, AutoCO achieves $T_{tff}$ values as low as 6.39 minutes for small instances and 11.65 minutes for large instances. This indicates that it generates feasible solutions significantly faster than baseline methods, with FunSearch ranging from 19.73 to 40.29 minutes, EoH ranging from 8.67 to 14.23 minutes, and ReEvo achieving 12.73 minutes for small instances and 12.54 minutes for large instances. Furthermore, AutoCO reduces $T_{stag}$ to a minimum of 5.99 minutes, compared to 8.71 to 28.49 minutes for FunSearch, 19.53 minutes for EoH, and 9.27 to 24.46 minutes for ReEvo, enhancing overall runtime efficiency.

    \item \textbf{Computational Time Distribution}: Figure \ref{fig:time} illustrates a comprehensive breakdown of AutoCO's time allocation. The pie chart demonstrates that our computational effort is distributed efficiently among global exploration, solution generation, constraint verification, and evolutionary refinement, instead of being concentrated in any single stage. This balanced distribution underscores AutoCO's holistic optimization strategy.
\end{itemize}

\begin{table}[tb]
\centering
\caption{Temporal performance characterization. (minutes)}
\label{tab:time_mdd}
\scriptsize
\setlength{\tabcolsep}{1.8pt}
\renewcommand{\arraystretch}{1.2} 
\begin{tabular}{@{}l*{9}{c}@{}}
\toprule
\multirow[c]{2}{*}{\textbf{Method}} & 
\multicolumn{3}{c}{\boldmath$T_{e2e}$↓} & 
\multicolumn{3}{c}{\boldmath$T_{tff}$↓} & 
\multicolumn{3}{c}{\boldmath{$T_{Stag}$↓}} \\
\cmidrule(lr){2-4} \cmidrule(lr){5-7} \cmidrule(l){8-10}
 & \textbf{S} & \textbf{M} & \textbf{L} & \textbf{S} & \textbf{M} & \textbf{L} & \textbf{S} & \textbf{M} & \textbf{L} \\
\midrule
FunSearch & 53.03 & 49.54 & 137.93 & 40.29 & 31.73 & 19.73 & \textbf{8.71} & 10.27 & 28.49 \\
EoH & 35.35 & 32.11 & 94.64 & 14.21 & 8.67 & 14.23 & 14.79 & 21.33 & 19.53 \\
ReEvo & \textbf{30.21} & \textbf{29.93} & \textbf{86.97} & 12.73 & \textbf{7.02} & 12.54 & 9.27 & 19.98 & 24.46 \\
\midrule
AutoCO & 65.07 & 104.11 & 108.87 & \textbf{6.39} & 9.31 & \textbf{11.65} & 25.53 & \textbf{5.99} & \textbf{11.77} \\
\bottomrule
\end{tabular}
\end{table}

Although AutoCO's end-to-end runtime appears longer (ranging from 65.07 to 108.87 minutes), it is crucial to note that \textit{the structured time allocation allows for efficient feasible solution generation while managing low performance stagnation. This results in a more comprehensive exploration of the solution space. Compared to expert manual design, this runtime remains competitive and satisfactory.}

% ===== Ablation Studies on Core Mechanisms =====
\subsection{Ablation Study}  
\label{subsec:ablation}  

We quantify the contributions of each component through the optimality gap increment ($\Delta$gap) relative to the complete AutoCO method. By selectively removing the constraint relaxation module (w/o $\sigma$), three-step generation strategy (w/o 3-steps), MCTS mechanism (w/o MCTS), and bidirectional information exchange(w/o Bidirectional), we systematically evaluate the individual contributions of each module in the AutoCO method. As shown in Table~\ref{tab:ablation}, the ablation experiments reveal the critical roles of these components: the constraint relaxation module causes an average performance degradation of +21.55\%, indicating its crucial role in strategy formulation; the three-step generation strategy leads to a +14.10\% performance decline, demonstrating the importance of structured generation; the MCTS mechanism shows a +17.46\% performance drop, highlighting the significance of global search in avoiding local optima; the bidirectional coevolution mechanism results in a +23.73\% performance degradation, showcasing its substantial contribution to overall optimization.\textit{The ablation experiments substantiate the meticulously designed modules and elucidate their synergistic optimization capabilities within AutoCO.}
\begin{table}[h]  
\centering  
\caption{Ablation results of different variants. ($\Delta$gap\%$\downarrow$)}  
\label{tab:ablation}  
\scriptsize
\setlength{\tabcolsep}{2.5pt}
\renewcommand{\arraystretch}{1.2} % 增加行高
\begin{tabular}{@{}lcccc@{}}  
\toprule  
\textbf{Variant} & \textbf{VRPTW} & \textbf{VRPTW-fuel} & \textbf{SFL} & \textbf{Avg.} \\  
\midrule  
\textbf{Full AutoCO} & \textbf{0.0} & \textbf{0.0} & \textbf{0.0} & \textbf{0.0} \\  
w/o $\sigma$ & +23.01 & +24.98 & +16.67 & +21.55 \\  
w/o \text{3-steps}  & +14.75 & +15.81 & +11.74 & +14.10 \\  
w/o  \text{MCTS } & +15.42 & +23.34 & +13.62 & +17.46 \\  
w/o  \text{Bidirectional} & +25.07 & +25.97 & +20.15 & +23.73 \\  
\bottomrule  
\end{tabular}
\end{table}

\subsection{Relaxation Strategy Effectiveness Validation}
\label{subsec:relaxation_validation}

To validate whether AutoCO's autonomous relaxation strategies outperform expert-designed alternatives, we apply a base search method across different feasible regions on SFL-8 and SFL-5(Dual), examining three relaxation strategies: (1) Original Constraints (Unrelaxed), (2) AutoCO-Designed Strategy $\sigma^*$, and (3) Expert-Designed Strategy (details in Appendix C), and assess their success rates under varying computational budgets.

\begin{figure}[htb]
\centering
\includegraphics[width=\columnwidth]{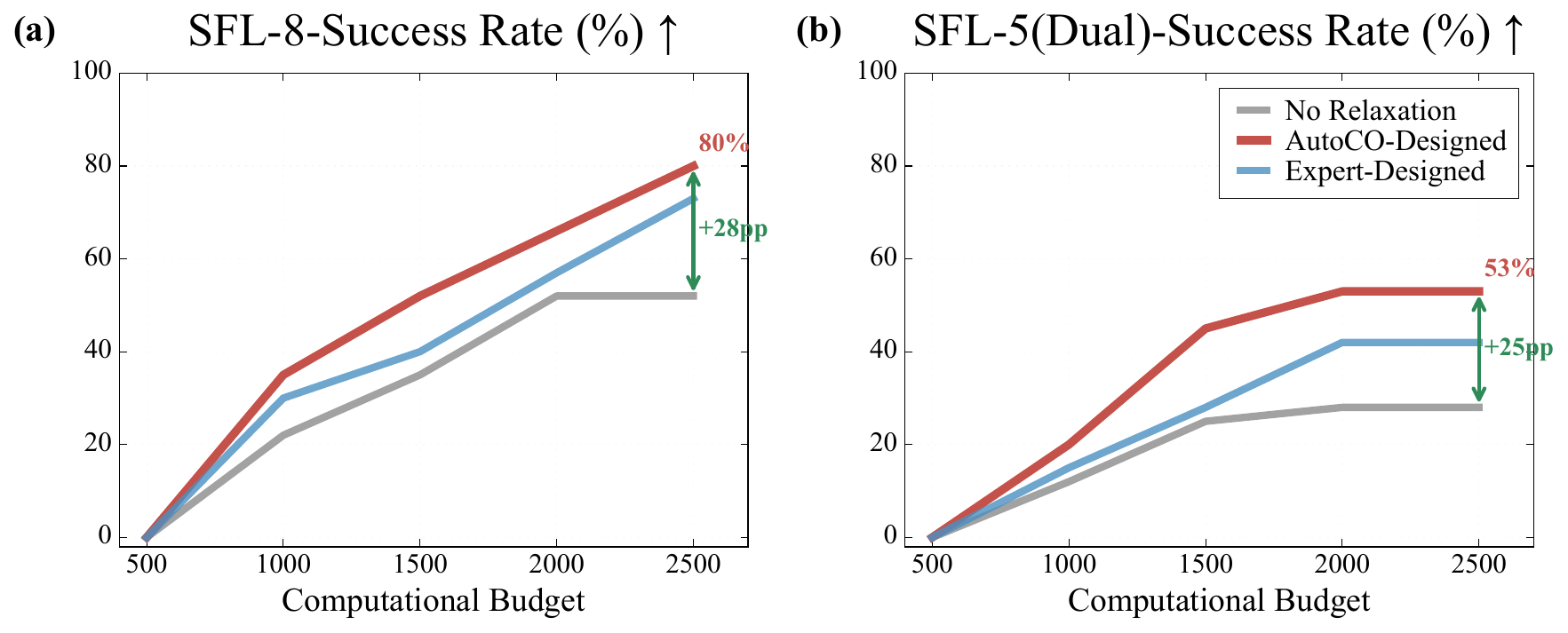}
\caption{Effectiveness of various constraint relaxation strategies under different computational budgets on SFL8 and SFL-5 (Dual).}
\label{fig:relaxation_budget_comparison}
\end{figure}

Figure~\ref{fig:relaxation_budget_comparison} demonstrates performance across computational budgets. On SFL-8, original constraints plateau at 52\%, while AutoCO strategy rapidly progresses from 0\% to 80\% success rate. AutoCO outperforms expert strategy, showing 1.1 times improvement at 2500 iterations. On SFL-5(Dual), which contains two safety regions, making the problem more challenging, performance gains are more modest, with AutoCO reaching 53\% success. \textit{AutoCO exhibits the most favorable ascending trend among all strategies, confirming that LLM-generated relaxation strategies effectively transform constraint navigation by constructing more accessible feasible regions than both rigid formulations and static expert designs.}

\subsection{Optimization Dynamics Analysis of LLM-based methods}
\label{subsec:dynamics}

\begin{figure}[htb]
\centering
\includegraphics[width=\columnwidth]{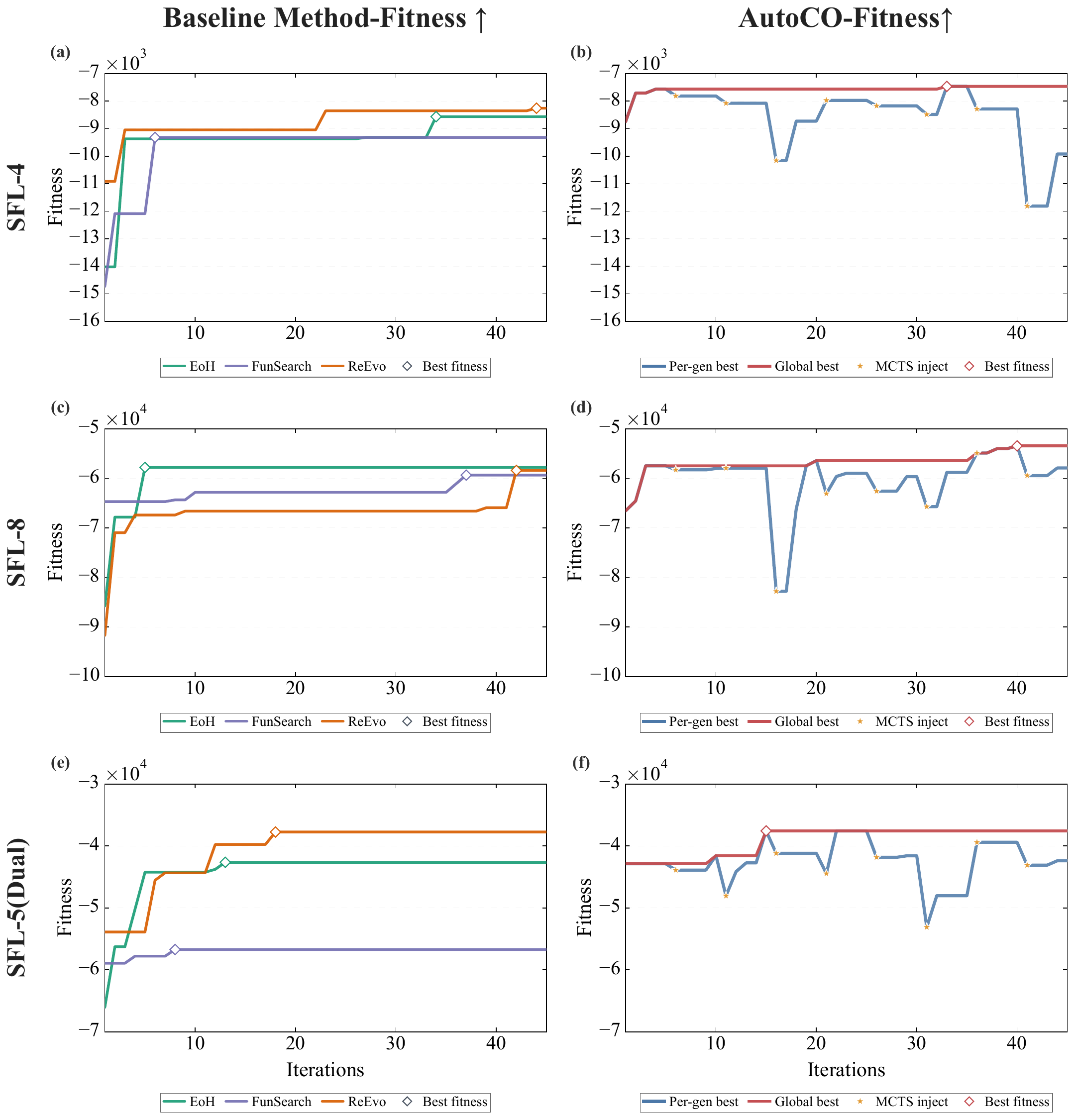}
\caption{Convergence dynamics and fitness analysis of LLM-based methods across SFL datasets: current llm-based methods (subplots a, c, e) and AutoCO (subplots b, d, f).}
\label{fig:six_datasets_comparison}
\end{figure}

To systematically analyze the fundamental differences in optimization behaviors between traditional LLM-based methods and our AutoCO method, we conduct a comparative analysis across three SFL datasets using Figure \ref{fig:six_datasets_comparison}, comparing baseline methods and AutoCO convergence dynamics.

Current LLM-based methods exhibit clear stagnation: After initial solutions, they focus exclusively on local optimization. In SFL-4, FunSearch converges at generation 5, while ReEvo and EoH attempt performance improvements by fine-tuning solutions between generations 35-40, remaining trapped in limited local search spaces.

AutoCO reveals distinctly different dynamics. Despite fitness stagnation, it maintains active search through bidirectional exchange and MCTS global strategies. As shown in subfigures b and d, AutoCO endures prolonged stagnation but continuously probes the solution space by injecting global information, ultimately breaking through initial constraints and achieving significant fitness gains.

Empirical results clearly validate AutoCO's key advantages: when traditional methods permanently stagnate after initial convergence, \textit{AutoCO systematically escapes local optima by injecting global strategies from MCTS, temporarily accepting suboptimal solutions to accumulate global information.} Although the experiment, constrained by time limitations, is set to 45 generations, the convergence plot demonstrates that AutoCO continues global exploration in later stages and achieves effective improvements, suggesting that with sufficient computational resources, AutoCO could potentially collect more comprehensive global information and achieve further performance enhancements.

\section{Conclusion and Future Work}
\label{sec:conclusion}

This paper introduces AutoCO, an end-to-end method for automated constraint optimization for COPs that transforms LLMs from passive validators into proactive strategy architects. Our key innovation lies in the systematic integration of three complementary components: (1) a structured three-step constraint relaxation method that codifies expert knowledge into LLM-interpretable principles, (2) a novel triple-representation scheme synchronizing constraint strategies, algorithmic concepts, and executable code across abstraction levels, and (3) a bidirectional coevolution mechanism that synergistically combines EA's local refinement with MCTS's global exploration to navigate solution spaces. Extensive experiments on VRPTW, SFL, and VRPTW-fuel verify AutoCO's effectiveness and generalizability. While AutoCO represents a significant advance in autonomous optimization of COPs, future research could focus on: (1) extending multi-objective optimization with more sophisticated constraint coupling strategies, and (2) developing distributed coevolution frameworks for scalable optimization processes.

\clearpage

\section*{Limitations}
This work primarily focuses on solving COPs through LLM-driven constraint relaxation and bidirectional coevolution. While the approach effectively handles complex constraints in domains like vehicle routing and facility layout, it overlooks several important aspects that warrant consideration. 

First, the methodological scope, though validated on three challenging benchmarks, remains limited to static optimization problems with deterministic constraints, leaving stochastic and fully dynamic optimization environments unexplored. Additionally, as problem dimensionality increases, the strategy space grows exponentially, potentially affecting the scalability of the coevolution mechanism.

Second, while AutoCO demonstrates efficiency in generating initial feasible solutions, the computational overhead from maintaining multiple synchronized components, including LLM reasoning, evolutionary algorithms, and Monte Carlo tree search—results in longer end-to-end runtime compared to some baseline methods. This quality-efficiency trade-off may limit applicability in scenarios requiring real-time decision-making.

Furthermore, the method's performance is influenced by the reasoning and coding capabilities of LLMs. While the triple-representation scheme provides structured guidance for constraint relaxation and code generation, the effectiveness of the initial strategy design and evolutionary refinement may vary with the LLM's analytical consistency, particularly during the early exploration stages.

Finally, while the empirical results strongly support the effectiveness of the bidirectional coevolution mechanism, its theoretical convergence properties remain to be fully established.

\section*{Ethics Statement}
The benchmarks and problem instances used in this work are all publicly available, ensuring no ethical concerns regarding data provenance. Beyond data sources, this work also considers the broader ethical implications of the AutoCO method.

First, regarding algorithmic robustness and reliability risks. As noted in the Limitations section, the method's performance is influenced by the reasoning capabilities of LLMs. Inconsistent analytical depth or programming proficiency in LLMs could affect the stability of constraint relaxation strategy generation, particularly during early exploration phases. In safety-critical applications such as urban logistics or facility planning, suboptimal solutions generated due to such instabilities could lead to operational inefficiencies or resource misallocation. This risk necessitates maintaining human oversight when deploying such automated systems in real-world scenarios.

Second, regarding potential misuse and applicability boundaries. The method's capability to automatically design optimization strategies for complex constraint problems raises considerations about its appropriate application domains. While developed for beneficial purposes in logistics and planning, similar techniques could potentially be applied to contexts with significant societal implications, such as resource allocation in critical infrastructure or strategic planning in sensitive domains. Domain-specific risk assessments and ethical reviews are strongly recommended before deploying such automated optimization systems, particularly in high-stakes environments where suboptimal decisions could have substantial consequences.

Finally, regarding computational resource and environmental considerations. The bidirectional coevolution mechanism achieves strong performance at the cost of substantial computational resources due to its multiple synchronized components. To responsibly manage this demand, strict time limits are enforced during experiments, and algorithmic optimizations are implemented to maximize search efficiency within these constraints. While computationally intensive, this investment is justified by the framework's capability to automate complex optimization design tasks that traditionally require extensive human expertise.

\bibliography{main}

\appendix

% \section{Appendix}
\label{sec:appendix}

\clearpage
\section{Details of The AutoCO Approach}
 In this part, we elaborate on the details of our proposed AotoCO (\textbf{Auto}mated \textbf{C}onstraint \textbf{O}ptimization via LLM-driven bidirectional coevolution) as well as experimental results.
\subsection{Triple Representation Scheme}
In Constraint Optimization Problems (COPs), constraint handling is a critical and complex challenge. Existing LLM-based optimization methods \cite{liuEvolutionHeuristicsEfficient2024a} are often limited to dual representations of algorithms and code, struggling to effectively generate feasible solutions.

\begin{figure}[htb]
\centering
\includegraphics[width=\columnwidth]{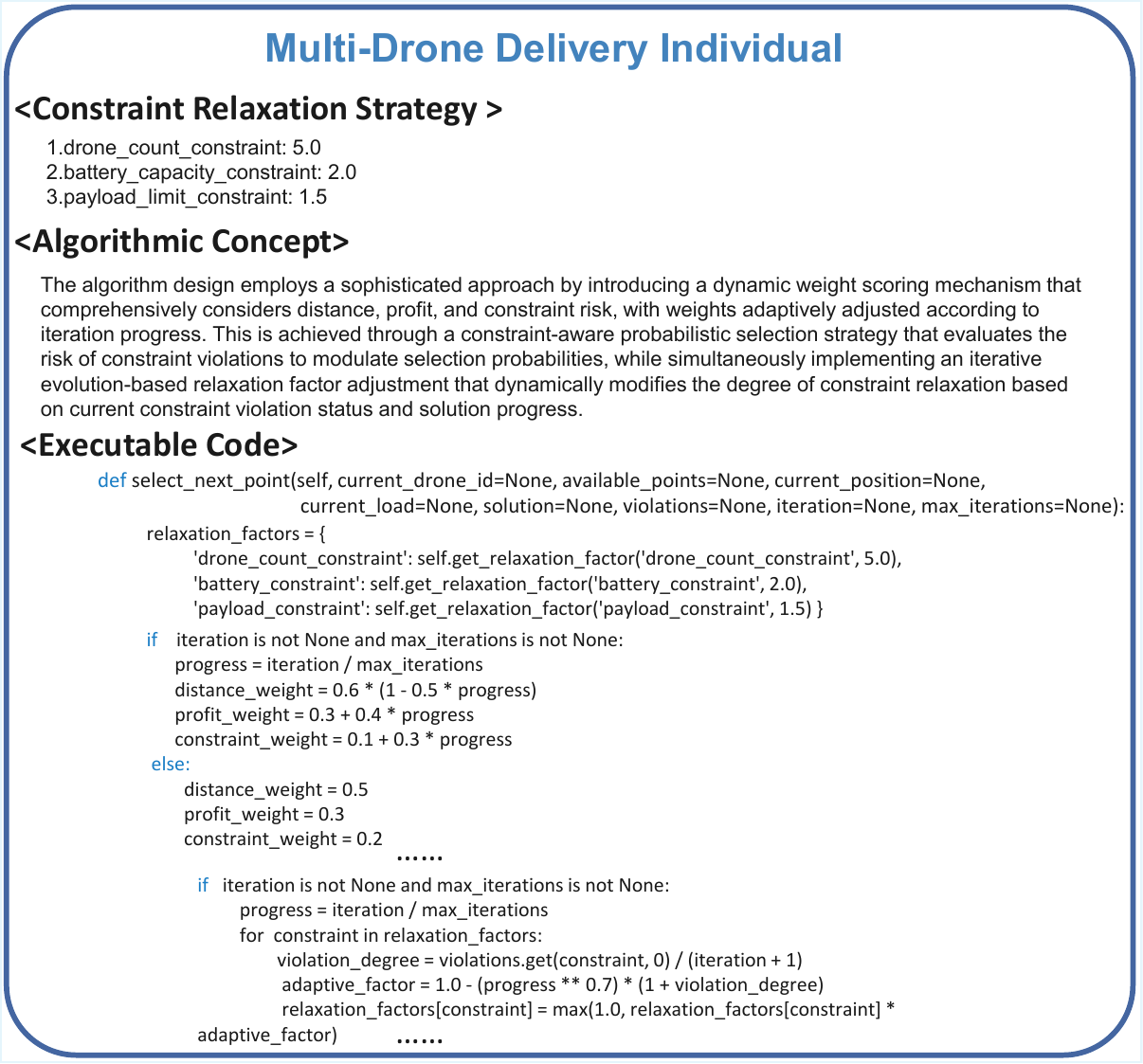}
\caption{The triple individual representation example for the multi-drone delivery problem.}
\label{fig:triple_representation}
\end{figure}

To overcome this limitation, we propose an innovative triple individual representation $\mathcal{I}_j = \langle \sigma_j, A_j, C_j \rangle$, focusing on systematically integrating constraint relaxation strategies. This representation, through explicitly introducing the constraint relaxation strategy $\sigma_j$, and containing algorithmic concept $A_j$ and executable code $C_j$, provides support for collaborative evolution in subsequent optimization stages.

%Figure \ref{fig:triple_representation} illustrates a triple individual representation example for the multi-drone delivery problem. \textcolor{red}{based on the figure, some more description?}

Figure \ref{fig:triple_representation} illustrates a triple individual representation example for the multi-drone delivery problem. As shown, the individual ${I}_j$ explicitly integrates three key components: 1) Constraint Relaxation Strategy ($\sigma_j$): Contains LLM-analyzed relaxation factors for critical constraints in multi-drone delivery (e.g., drone count: 5.0, battery: 2.0). 2) Algorithmic Concept ($A_j$): Describes the algorithmic implementation logic designed by LLM for constraint relaxation strategies, such as a dynamic weight scoring mechanism balancing distance, profit, and constraint risk. 3) Executable Code ($C_j$): The executable Python code designed by the LLM that integrates the relaxation strategy with the algorithmic concept.

\begin{algorithm}[tb]
\caption{Bidirectional Coevolution Mechanism }
\label{alg:strategy_search}
\textbf{Input}: Initial population $P_0 = \{\mathcal{I}_i = \langle \sigma_i, A_i, C_i \rangle\}$, problem instances $\Pi$, MCTS tree $\mathcal{T}$ \\
\textbf{Output}: $\mathcal{I}^*$
\begin{algorithmic}[1]
\While{not converged}
    \State \textbf{// Local Evolutionary Optimization}
    \State $P_{\text{off}} \gets \emptyset$
    \For{$\mathcal{I}_p \in \text{Select}(P_t)$} \Comment{Roulette: $P(i) \propto f(\mathcal{I}_i)$}
        \State $\mathcal{I}_c \gets \text{LLM\_Generate}(\mathcal{I}_p)$ \Comment{$\langle\sigma,A,C\rangle$}
        \State $f(\mathcal{I}_c) \gets \text{Eval}(C_c, \Pi)$ \Comment{Code Execution}
        \State $P_{\text{off}} \gets P_{\text{off}} \cup \{\mathcal{I}_c\}$
    \EndFor
    \State $P_t \gets \text{Top}_N(P_t \cup P_{\text{off}})$ \Comment{Elitism}
    
    \State \textbf{// Global MCTS Exploration}
    \For{$k=1$ \textbf{to} $K$}
        \State $v \gets \text{TreePolicy}(\mathcal{T})$ \Comment{UCT Selection}
        \State $R \gets f(\text{LLM\_Generate}(v)$ \Comment{Simulate $\sigma$}
        \State $\text{Backup}(v, R)$ \Comment{$Q_v \mathrel{+}= R, N_v \mathrel{+}= 1$}
    \EndFor
    \State $\sigma^* \gets \arg\max_{\sigma} Q_{\text{leaf}}/N_{\text{leaf}}$
    
    \State \textbf{// Bidirectional Transfer}
    \For{$\mathcal{I} \in P_t$}
        \State $\text{UpdateTree}(\mathcal{T}, \sigma_{\mathcal{I}}, f(\mathcal{I}))$ \Comment{Local→Global}
    \EndFor
    \If{$\text{Stagnant}(P_t)$} \Comment{$S_{\text{total}} < \theta$}
        \State $P_t \gets P_t \cup \langle \sigma^*, A, C \rangle$ \Comment{Global→Local}
    \EndIf
\EndWhile
\State \textbf{return} $\arg\max f(\mathcal{I})$
\end{algorithmic}
\end{algorithm}

\subsection{Bidirectional Coevolution Mechanism}
We devise a bidirectional coevolution mechanism combining a local Evolutionary Algorithm (EA) with a global Monte Carlo Tree Search (MCTS) to address constraint challenges in COPs. This mechanism enables effective navigation and optimization across diverse solution regions by leveraging MCTS’s global perspective to help EA escape local optima while enhancing resource efficiency through integrating EA’s exploration results into
MCTS. It comprises three key components: (1) local coevolutionary optimization, (2) global strategy exploration, and (3) bidirectional information exchange. The detail of the mechanism is shown in Algorithm \ref{alg:strategy_search}.

%The fitness evaluation process $f(\mathcal{I}) = \frac{1}{|\Pi|}\sum_{i=1}^{|\Pi|} f_i(\mathcal{I})$ is performed by executing code $C$ on problem instances $\pi_i \in \Pi$, where lower objective function values indicate better solutions. Upon detection of syntax or runtime errors, an LLM-based iterative repair mechanism is triggered (maximum $R_{\max}=3$ attempts), leveraging error messages, the algorithmic concept $A$, and problem specifications to generate corrected code $C'$. If repair attempts fail to achieve convergence, a penalty fitness $f_{\text{penalty}} = \varepsilon \cdot \min_j f(\mathcal{I}_j)$ (where $\varepsilon=0.01$) is assigned, ensuring $f_{\text{penalty}} \ll \min \{ f(\mathcal{I}_k) \mid \mathcal{I}_k \in P_{\text{valid}} \}$. The resulting fitness value $f(\mathcal{I})$ drives EA selection and MCTS reward propagation, thereby closing the optimization feedback loop.

\begin{algorithm}[H]
\caption{Code Execution and Individual Evaluation}
\label{alg:code_execution}
\begin{algorithmic}[1]
\Require Strategy $\sigma \in \Sigma$, Problem Instances $\Pi$
\Ensure Fitness Value $f(sigma) \in {R}^+$

\Comment{$\Sigma$: Strategy Space, $P$: Problem Instance Set}
\Comment{$R_{\max}$: Maximum Repair Attempts, Typically $R_{\max} = 3$}

\State $r \gets 0$ \Comment{Repair attempt counter}
\State $R_{\max} \gets 3$

\While{$r \leq R_{\max}$}
    \State $\xi \gets \mathcal{E}(\sigma, \Pi)$ \Comment{Execute strategy $\sigma$ on instances $\Pi$}
    \State $\mu \gets \omega(\xi)$ \Comment{Compute solution metrics}
    \State $\Phi \gets \chi(\xi)$ \Comment{Check constraint violations}
    
    \If{$\neg \Phi$} \Comment{No constraint violations}
        \State \Return $f(\sigma) \gets \lambda(\mu)$ \Comment{Calculate fitness from metrics}
    \Else
        \State $\sigma \gets \gamma(\sigma, \tau)$ \Comment{Repair strategy using LLM}
        \State $r \gets r + 1$
    \EndIf
\EndWhile
\State \Return $f_{\text{penalty}}$ \Comment{Penalty for repeated failures}
\end{algorithmic}
\end{algorithm}

\subsection{Code execution}

To ensure secure, reliable execution and robust evaluation of LLM-generated code $C$, AutoCO implements a lightweight isolated execution environment with comprehensive safety mechanisms. Resource exhaustion is prevented by enforcing strict execution time limits ($T_{\max}=60$s) and memory constraints ($M_{\max}=1024$MB).

The fitness evaluation process can be formulated as:

\begin{equation}
\begin{aligned}
f(\mathcal{I}) = \frac{1}{|\Pi|}\sum_{i=1}^{|\Pi|} f_i(\mathcal{I}),
\end{aligned}
\end{equation}
where $f(\mathcal{I})$ is the average fitness of an individual, $|\Pi|$ represents the total number of problem instances, and higher objective function values indicate better solutions.

The detailed process of code execution and individual evaluation is illustrated in Algorithm \ref{alg:code_execution}. Upon detection of syntax or runtime errors, an LLM-based iterative repair mechanism is triggered (maximum attempts $R_{\max}$). \textbf{If no feasible solution is found or execution fails after reaching the repair limit, the individual is assigned an infinite fitness value.} The detail is shown in Algorithm \ref{alg:code_execution}.

\section{Prompt Design}

\subsection{LLM-Driven Constraint Relaxation Strategy Generation}
In the process of constraint relaxation strategy generation for COPs, we systematically design three key prompt stages, aiming to fully utilize the reasoning capabilities of LLMs to transform constraint handling from experience-driven to intelligence-driven methods. Each prompt was carefully crafted to guide LLM to conduct in-depth and systematic constraint analysis from different perspectives.

\subsubsection{Constraint Importance Analysis Prompt}

\begin{figure}[htbp]
\centering
\includegraphics[width=\columnwidth]{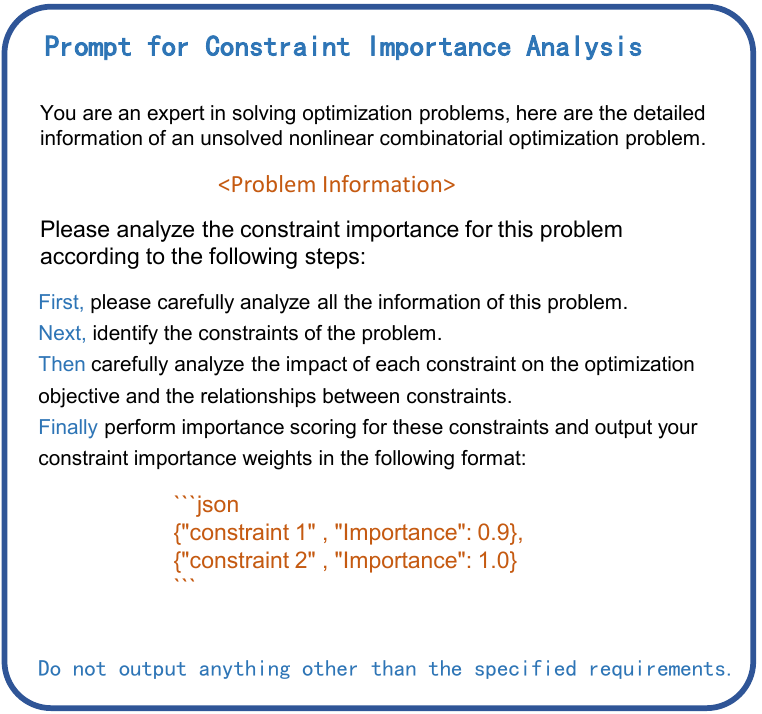}
\caption{Constraint importance analysis prompt}
\label{fig:constraint_importance_prompt}
\end{figure}

In the first stage, we guide the LLM to assess the importance of constraints through a carefully designed prompt. Figure \ref{fig:constraint_importance_prompt} demonstrates the specific prompt design.

The core objective of this prompt is to transform complex constraint parsing into a structured importance assessment, enabling the LLM to intelligently identify and weigh the criticality of various constraints from the problem description.

\subsubsection{Constraint Relaxation Range Suggestion Prompt}

\begin{figure}[htbp]
\centering
\includegraphics[width=\columnwidth]{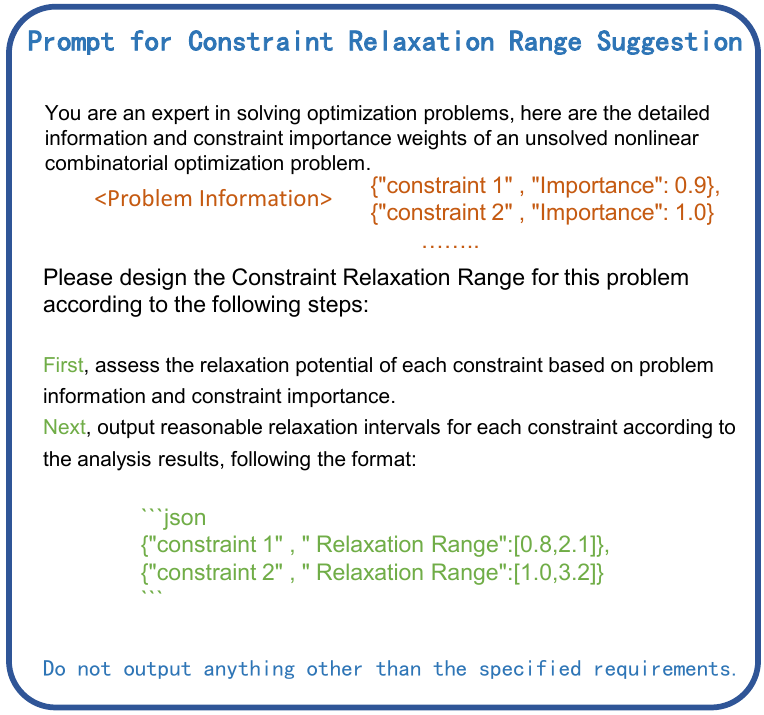}
\caption{Constraint relaxation range suggestion prompt}
\label{fig:constraint_relaxation_range_prompt}
\end{figure}

Based on the constraint importance analysis in the first stage, the second stage prompt (as shown in Figure \ref{fig:constraint_relaxation_range_prompt}) aims to determine reasonable relaxation ranges for each constraint.

This prompt guides the LLM to intelligently recommend minimum and maximum relaxation factors based on the constraint importance and problem characteristics, laying the foundation for subsequent strategy generation.

\subsubsection{Constraint Relaxation Strategy Generation Prompt}

\begin{figure}[h]
\centering
\includegraphics[width=\columnwidth]{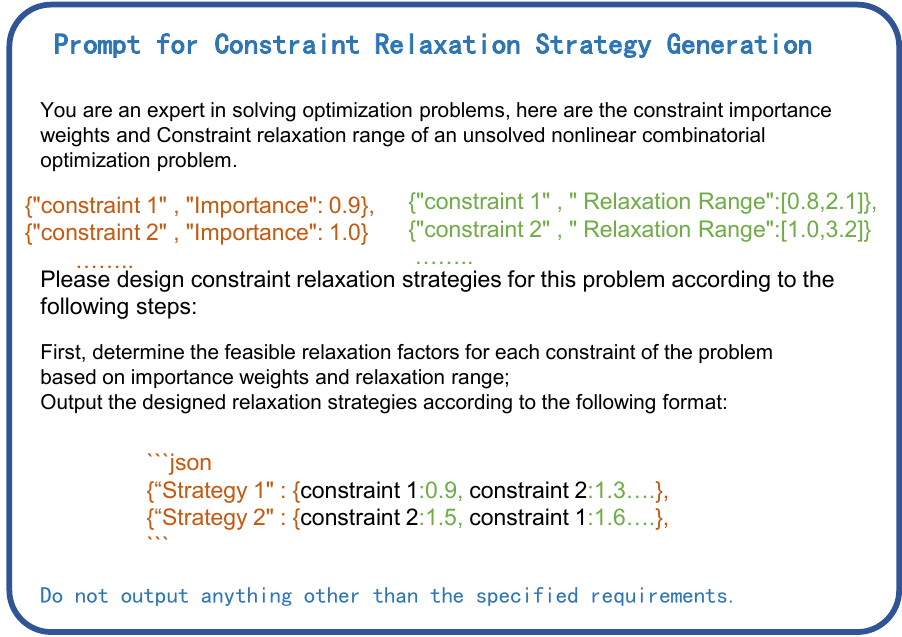}
\caption{Constraint relaxation strategy generation prompt}
\label{fig:constraint_relaxation_strategy_prompt}
\end{figure}

Finally, the prompt shown in Figure \ref{fig:constraint_relaxation_strategy_prompt} integrates the analysis results from the previous two stages to generate a series of diverse and meaningful constraint relaxation strategies.

Through this systematic prompt design, we successfully transformed the constraint handling method from expert experience to an LLM-based automated strategy generation approach. This not only enhances the intelligence of constraint handling but also significantly reduces dependence on domain experts.

\subsection{LLM-Guided Evolutionary Strategy Prompt Design}
We introduce an LLM-guided coordinated evolution to enhance the EA's exploration capabilities and achieve strategy-code-concept coevolution, ultimately realizing the local coevolutionary optimization part of our bidirectional coevolution mechanism. Specifically, through carefully designed prompt strategies, the LLM guides EA mutation and crossover across three levels: constraint relaxation strategies, algorithmic concepts, and executable code, transforming traditional evolutionary search into an intelligently adaptive exploration system, with the following prompt design details.

\begin{figure}[h]
\centering
\includegraphics[width=\columnwidth]{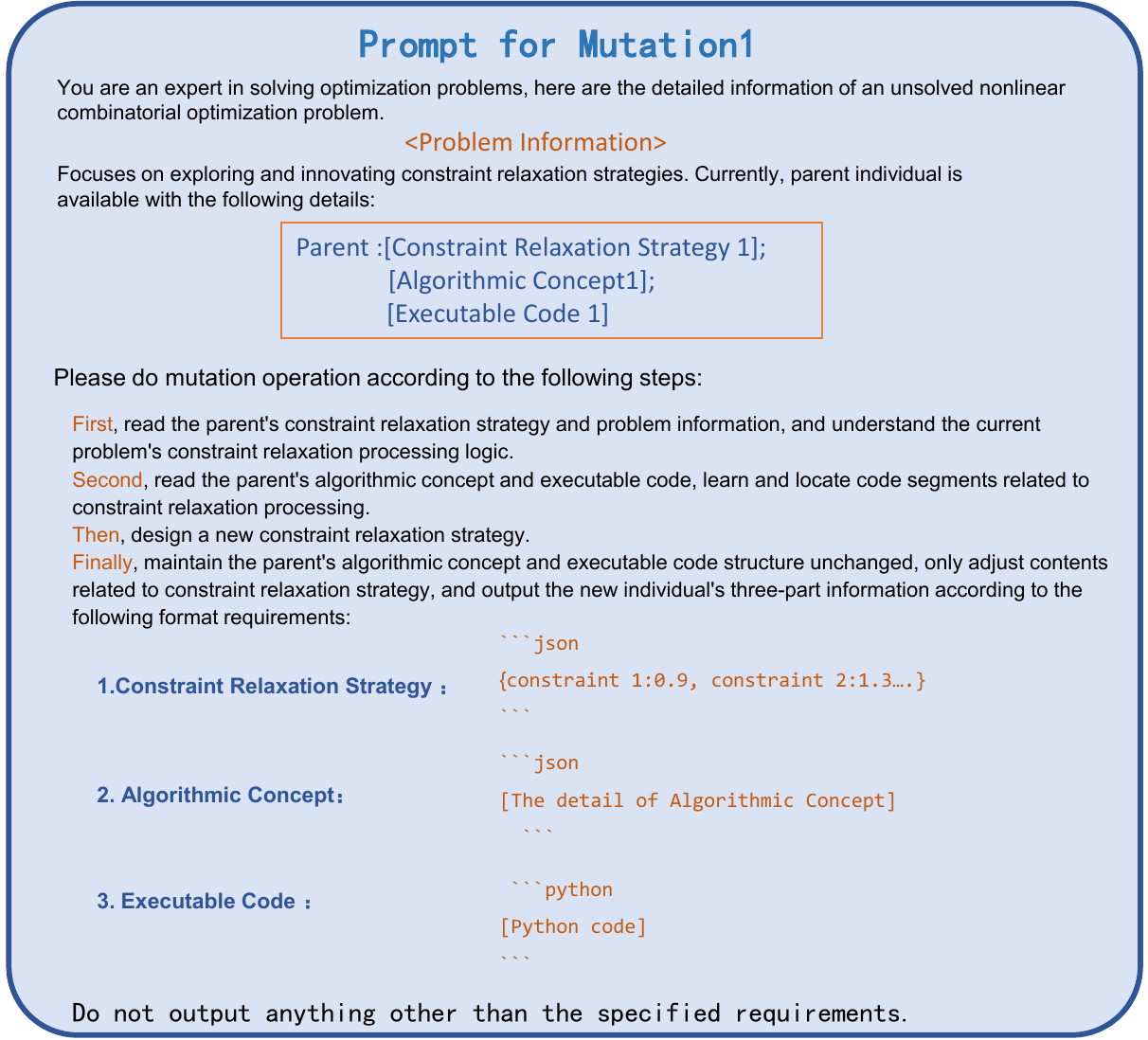}
\caption{The prompt detail of mutation strategy 1}
\label{fig:mutation_strategy1}
\end{figure}

\subsubsection{Mutation Strategy Prompts}
Mutation strategies are critical for exploring algorithm solution spaces. We design three mutation strategy prompts across different dimensions, aiming to drive autonomous innovation in algorithmic concept, constraint relaxation strategies, and executable code through structured and targeted guidance.

\paragraph{Mutation Strategy 1: Constraint Relaxation Strategies Prompt}
The Constraint Relaxation Prompt focuses on guiding the LLM to conduct an in-depth analysis of existing constraint relaxation strategies and problem information, redesigning constraint relaxation strategies while maintaining the unchanged algorithm concept and code structure, thereby achieving flexible adjustment and optimization of constraint strategies (as shown in Figure \ref{fig:mutation_strategy1}).

\begin{figure}[h]
\centering
\includegraphics[width=\columnwidth]{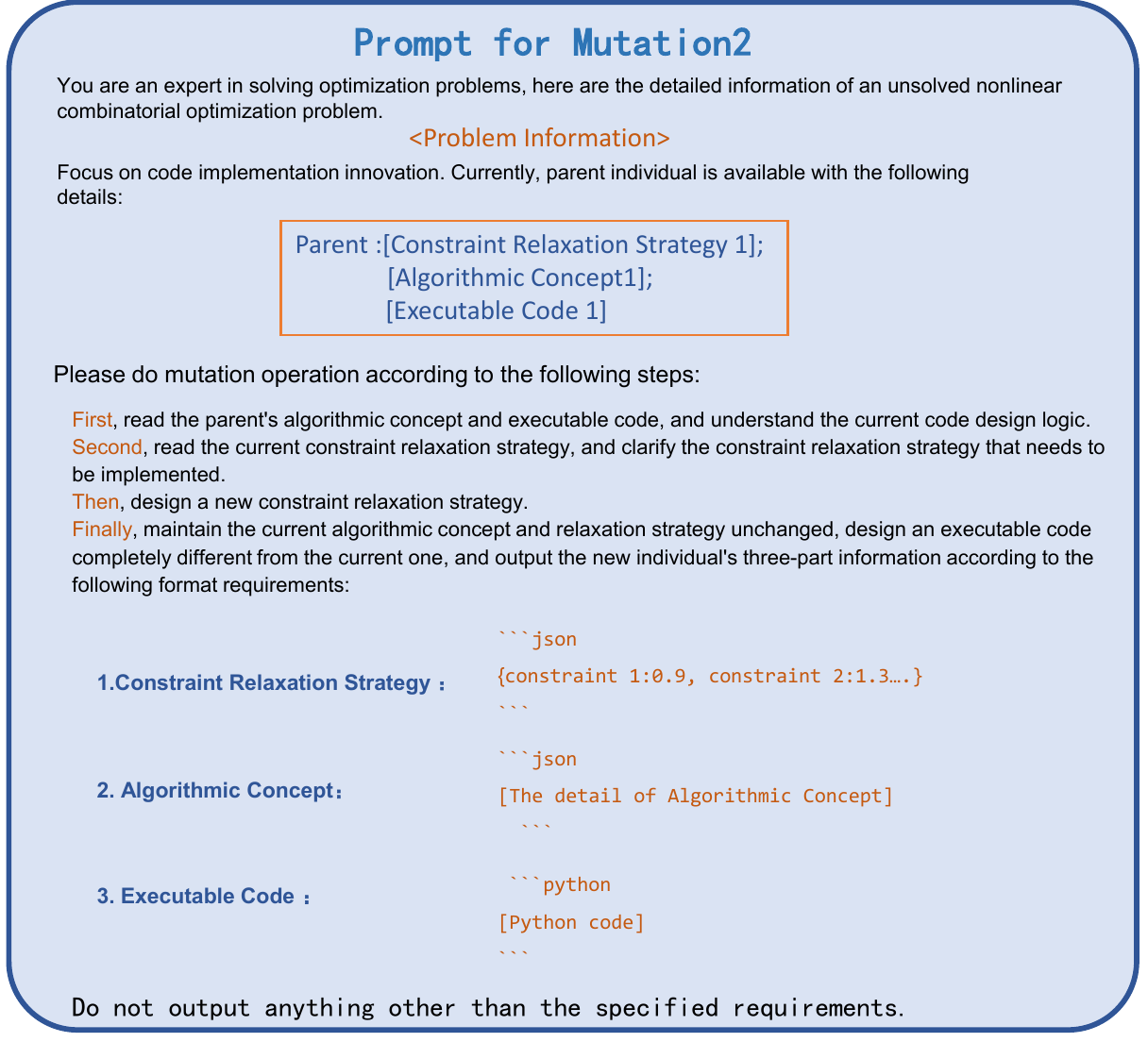}
\caption{The prompt detail of mutation strategy 2}
\label{fig:mutation_strategy2}
\end{figure}

\begin{figure}[h]
\centering
\includegraphics[width=\columnwidth]{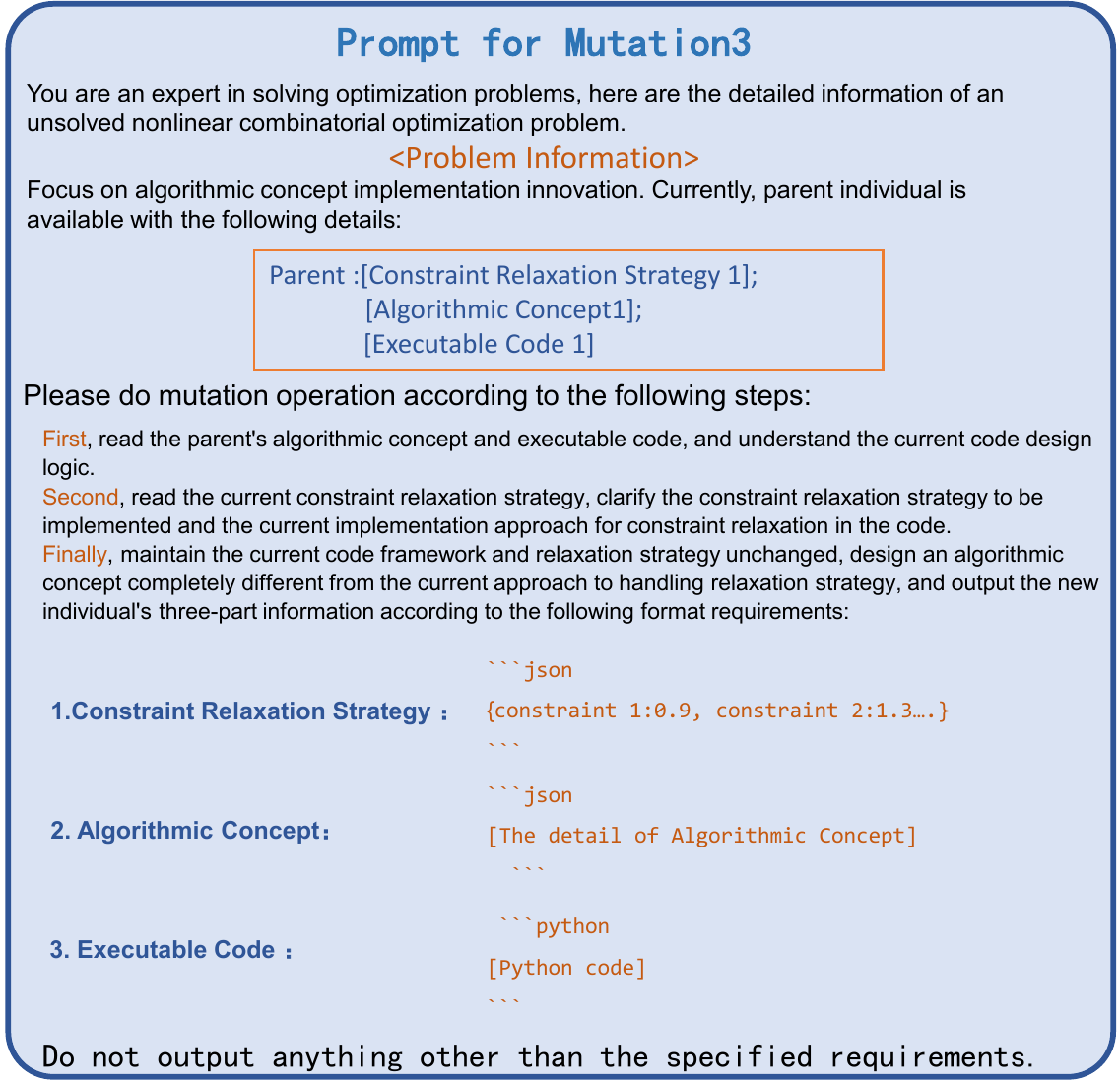}
\caption{The prompt detail of mutation strategy 3 }
\label{fig:mutation_strategy3}
\end{figure}
\paragraph{Mutation Strategy 2: Code Implementation Prompt }
By delving into the subtle nuances of current algorithm concepts and constraint relaxation strategies, the prompt is dedicated to guiding the LLM to discover novel implementation pathways, constructing fundamentally different executable code while preserving the original strategic framework (as shown in Figure \ref{fig:mutation_strategy2}).

\paragraph{Mutation Strategy 3: Algorithmic concepts Prompt}
Grounded in the existing code framework and constraint relaxation strategy, the prompt is committed to breaking through the current algorithmic concept, introducing a conceptually distinct approach, and guiding the LLM to reimagine the core mechanisms of problem-solving and strategy processing (as shown in Figure \ref{fig:mutation_strategy3}).

\subsubsection{Crossover Strategy Prompts}
\begin{figure}[h]
\centering
\includegraphics[width=\columnwidth]{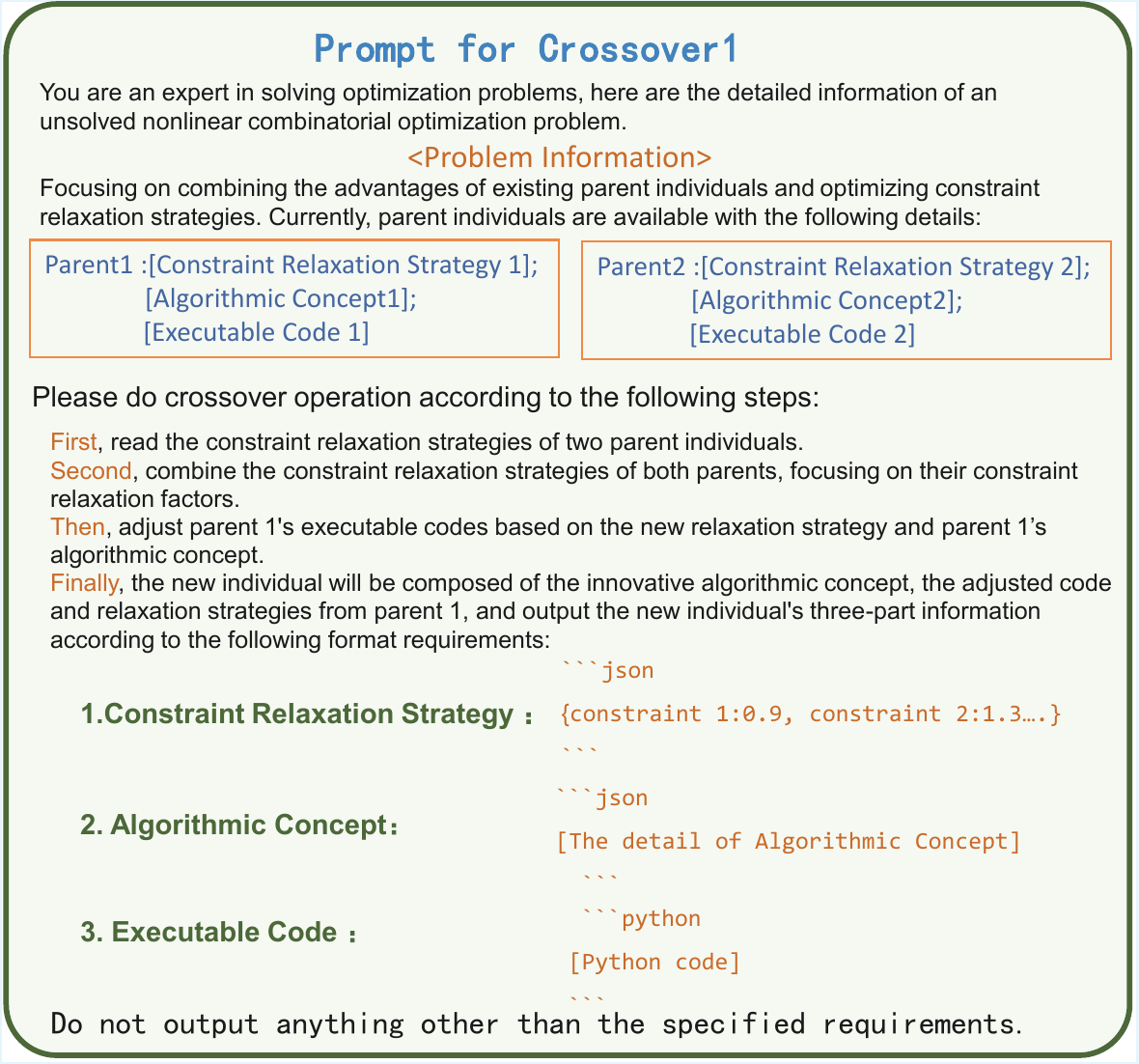}
\caption{The prompt detail of crossover strategy 1}
\label{fig:crossover_strategy1}
\end{figure}

Crossover strategies generate more promising new individuals by integrating advantages in constraint relaxation strategies, algorithmic concepts, or implementation code across different parent algorithms. We design three crossover strategy prompts to embody the coevolutionary optimization.
\paragraph{Crossover Strategy 1: Constraint Relaxation Strategies Prompt}This crossover prompt focuses on integrating and optimizing constraint relaxation strategies across parent algorithms. By systematically combining and analyzing constraint relaxation factors, exploring their interactions, and adjusting executable codes based on innovative relaxation approaches, this prompt aims to generate offspring with enhanced constraint processing capabilities, with primary emphasis on refining relaxation mechanisms (as shown in Figure \ref{fig:crossover_strategy1}).

\begin{figure}[h]
\centering
\includegraphics[width=\columnwidth]{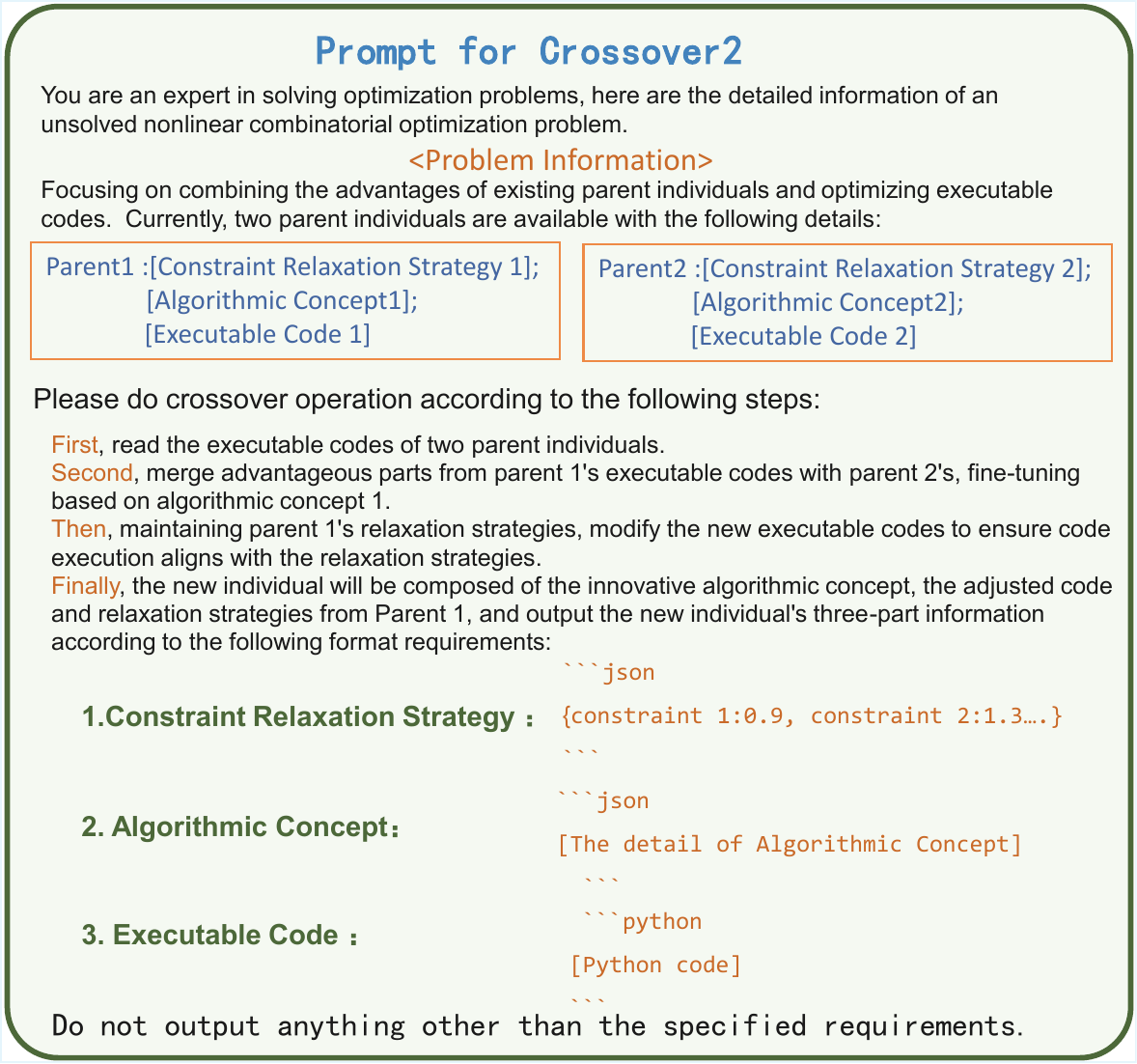}
\caption{The prompt detail of crossover strategy 2}
\label{fig:crossover_strategy2}
\end{figure}

\paragraph{Crossover Strategy 2: Code Implementation Prompt}The prompt aims to achieve algorithmic implementation optimization through strategic code merging and fine-tuning. By extracting advantageous code segments from parent algorithms, maintaining original relaxation strategies, and ensuring code execution alignment, this prompt enables the creation of more adaptive and sophisticated algorithmic implementations.(as shown in Figure \ref{fig:crossover_strategy2})

\begin{figure}[h]
\centering
\includegraphics[width=\columnwidth]{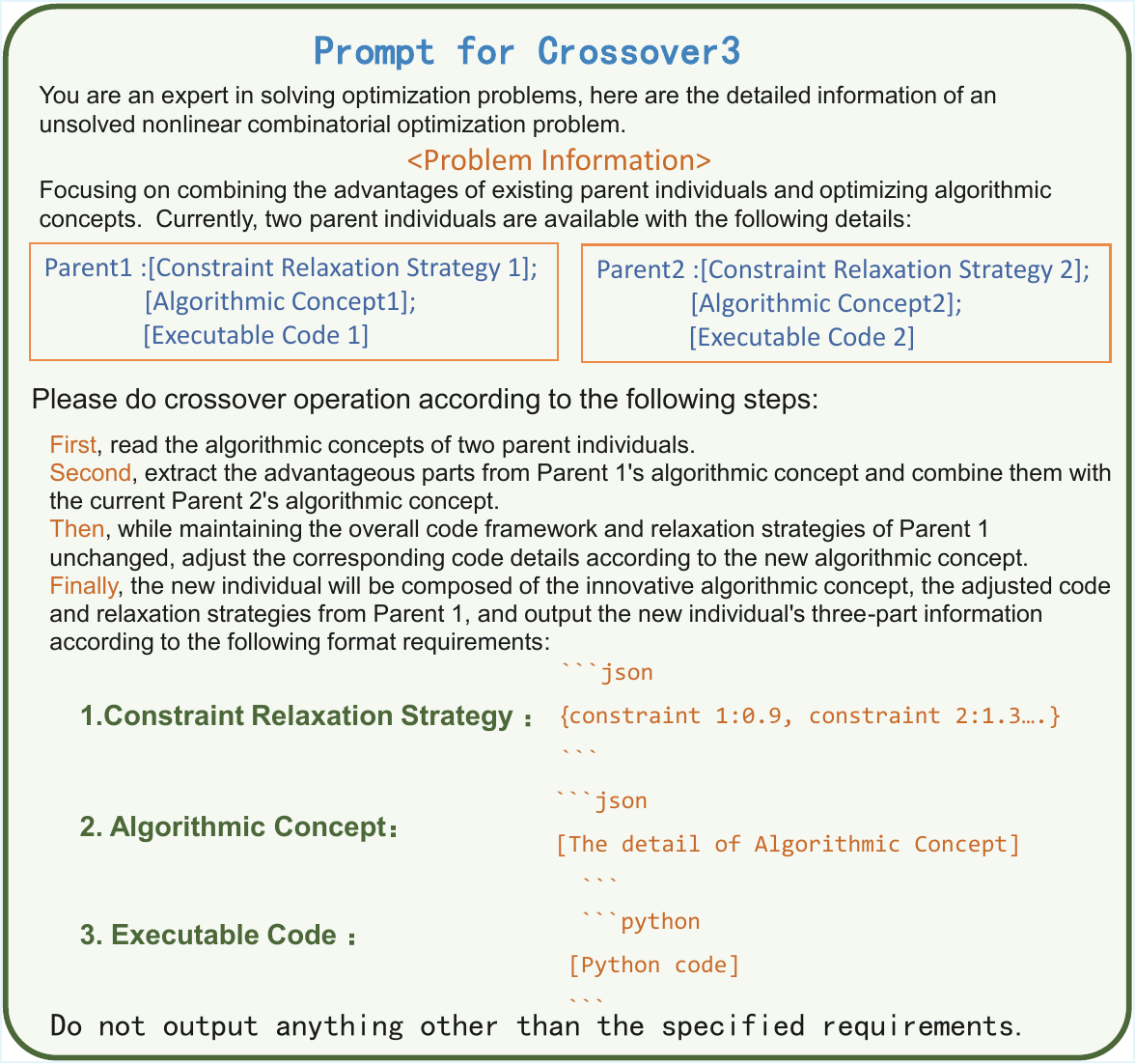}
\caption{The prompt detail of crossover strategy 3}
\label{fig:crossover_strategy3}
\end{figure}

\paragraph{Crossover Strategy 3: Algorithmic concepts Prompt}The prompt explores innovative integration of algorithmic core concepts while preserving the original code framework and relaxation strategies. By extracting advantageous conceptual elements from parent algorithms and carefully adjusting code details, this prompt facilitates the development of conceptually enriched algorithm variants that maintain the essential structural integrity of the original algorithms (as shown in Figure \ref{fig:crossover_strategy3}).

\section{Experimental Details}
\subsection{Benchmarks}
To validate AutoCO's generalizability, we strategically select three COPs that systematically span the key challenge dimensions: Vehicle Routing Problem with Time Windows (VRPTW), Safety Facility Layout (SFL), and Vehicle Routing Problem with Time Windows and fuel (VRPTW-fuel).

\paragraph{Safety Facility Layout}
The SFL originates from the CMU-IBM Open Source MINLP Project: \url{https://egon.cheme.cmu.edu/ibm/page.htm}. The problem focuses on optimizing facility positioning within a given area, requiring simultaneous satisfaction of multiple complex spatial and safety constraints, and addressing significant interactions between constraints. Specifically, the Safety Facility Layout (SFL) problem minimizes the total cost ($C_{\text{total}} = \sum_{i < j} C_{ij} \cdot |\mathbf{c}_i - \mathbf{c}_j|2 + \sum{i} C_i \cdot |\mathbf{c}_i - \mathbf{c}_{i0}|_2 $), where ( $||\cdot\|_2$ ) represents Euclidean distance, subject to facility non-overlap restrictions (axis-aligned disjunction: $((x_i + w_i \le x_j) \lor (x_j + w_j \le x_i) \lor (y_i + h_i \le y_j) \lor (y_j + h_j \le y_i))$, positional range boundaries ( $[\underline{x}_i, \overline{x}_i] \times [\underline{y}_i, \overline{y}_i]$), safety area limits ( $| \mathbf{c}_i - \mathbf{o}_k |_2 \leq R_k$ ), and global boundary requirements. The problem introduces significant nonlinearity through Euclidean norms and circular safety inclusion constraints, creating complex, fragmented solution spaces, with test instances including SFL-4 (4 facilities, single safety zone), SFL-8 (8 facilities, single safety zone), and SFL-5 (5 facilities, dual safety zones), representing varying levels of geometric complexity and constraint interactions.

\paragraph{Vehicle Routing Problem with Time Windows}
The VRPTW seeks a set of vehicle routes starting and ending at a single depot (node $0$) to serve customers $N=\{1,\dots,n\}$ with travel distances $d_{ij}$, demands $q_i$, service times $s_i$, and time windows $[e_i,\ell_i]$. Each customer is visited exactly once; vehicle load may never exceed capacity $Q$; early arrival allows waiting; and exceeding $\ell_i$ produces lateness. A solution is encoded as a giant-tour permutation decoded into a route family $\mathcal{R}$. For a route $r\in\mathcal{R}$ with sequence
$(v_0 = 0,\; v_1,\ldots,v_m,\; v_{m+1}=0),$
its distance is $D_r = \sum_{k=0}^{m} d_{v_k v_{k+1}}$,$\qquad D = \sum_{r\in\mathcal{R}} D_r$.
Arrival times propagate (with $t_{v_0}=0$,$s_0=0$) by$t_{v_{k+1}} = \max\bigl(t_{v_k} + d_{v_k v_{k+1}},\, e_{v_{k+1}}\bigr) + s_{v_{k+1}}, \quad k=0,\dots,m .$
Lateness for customer $i$ is $\max(0, t_i - \ell_i)$. A penalized objective consistent with our implementation is$C_{\text{VRPTW}} \;=\; D
\;+\; \alpha_{\text{time}} \sum_{i\in N} \max(0, t_i - \ell_i)
\;+\; \alpha_{\text{fleet}} \max\bigl(0,\, |\mathcal{R}| - K_{\max}\bigr),$
where $K_{\max}$ bounds the fleet size. Constraints: (i) each $i\in N$ appears in exactly one route; (ii) every route starts/ends at the depot; (iii) cumulative load on any prefix does not exceed $Q$; (iv) time-window upper bounds $\ell_i$ respected (or penalized if violated); (v) optional fleet-cardinality limit $|\mathcal{R}|\le K_{\max}$. Classical Solomon instances \cite{solomon1987} are challenging due to the interdependence of temporal feasibility and capacity-driven splitting.

\paragraph{Vehicle Routing Problem with Time Windows and fuel}
VRPTW-Fuel augments VRPTW by adding a per-route fuel capacity constraint. Retaining base notation, each arc $(v_k, v_{k+1})$ on a route carries load-dependent fuel:$
f_{v_k v_{k+1}} = \bigl(\beta + \gamma L_{v_k}\bigr)\, d_{v_k v_{k+1}},$
with $L_{v_k}$ denoting remaining load just before traversing the arc. Route fuel $F_r = \sum_{k=0}^{m} f_{v_k v_{k+1}}$ must satisfy $F_r \le C_{\text{fuel}}$. Violations aggregate as $ V_{\text{fuel}} = \sum_{r\in\mathcal{R}} \max(0, F_r - C_{\text{fuel}}),$
yielding the penalized objective $C_{\text{VRPTW-Fuel}} = D + \alpha_{\text{time}} \sum_{i\in N} \max(0, t_i - \ell_i) + \alpha_{\text{fleet}} \max(0, |\mathcal{R}| - K_{\max}) + \alpha_{\text{fuel}} V_{\text{fuel}}.$
All classical constraints remain, with the added per-route fuel capacity. The affine load term $(\beta + \gamma L_{v_k})$ couples sequence, load depletion, and arc selection: front-loading high-demand customers reduces later fuel intensities but can degrade temporal feasibility or distance, sharpening trade-offs and fragmenting the search landscape. Ties among fully feasible solutions are further broken (implementation) by lower fleet size, then lower distance, then lower maximum single-route fuel consumption. 

\subsection{Experiment Implementation} All experiments use DeepSeek-R1 LLM (population size: 45, time limit: 2 hours). 100 independent runs per instance on Intel® Core™ i5-13400F/NVIDIA RTX 4060 Ti platform. Non-LLM baselines run as single-threaded CPU executables.

\subsection{Baseline and Baseline Setting}
To comprehensively evaluate AutoCO's performance, we selected diverse baseline methods spanning exact solvers, Reinforcement Learning, metaheuristics, and LLM-based approaches.

\subsubsection{Exact Solver}
\paragraph{Gurobi} As a leading commercial mixed-integer programming solver, Gurobi (see \url{https://www.gurobi.com/}) utilizes advanced mathematical programming techniques, such as Branch and Bound, Cut Generation, and preprocessing techniques. Its ability to precisely solve complex optimization problems and numerical stability make it the gold standard for exact solutions.

\subsubsection{Reinforcement Learning}
\paragraph{DeepACO} Introducing a neural-enhanced metaheuristic framework that generalizes and amplifies traditional Ant Colony Optimization (ACO) through deep reinforcement learning \cite{ye2023deepaco}. This approach employs a graph neural network to encode combinatorial problem instances, generating advanced heuristic priors that significantly accelerate the solution construction process. The framework is trained via proximal policy optimization, enabling adaptive pheromone initialization and dynamic parameter adjustment, which collectively enhance search efficiency and solution quality across diverse combinatorial problems.
\subsubsection{Metaheuristic Algorithms}
\paragraph{Simulated Annealing (SA)}Inspired by the physical annealing process, SA introduces a temperature parameter to control the probability of accepting suboptimal solutions \cite{SA1983}. At high temperatures, the algorithm is more inclined to accept suboptimal solutions, helping to escape local optima; as the temperature decreases, the algorithm gradually converges to the optimal solution.
\paragraph{Genetic Algorithm (GA)}Mimicking natural selection and evolutionary processes, GA optimizes solutions through population-level selection, crossover, and mutation operations \cite{GA1994}. Each generation selects the most optimal individuals through fitness evaluation, simulating the natural selection mechanism of biological evolution.
\paragraph{Particle Swarm Optimization (PSO)} Inspired by collective behaviors of bird flocks and fish schools, each "particle" in PSO represents a potential solution, exploring the solution space by mutual following and learning \cite{PSO1995}. Particles are influenced not only by their own historical optimal solutions but also guided by the global optimal solution.
\paragraph{Memetic Algorithm (MA)} Combining genetic algorithms' evolutionary mechanisms with local search intensification, MA implements a hybrid optimization paradigm \cite{MA1989}. This algorithm maintains population diversity through global evolutionary operations while employing problem-specific local search techniques to refine individual solutions. This dual-optimization strategy achieves an effective balance between exploration and exploitation, frequently demonstrating superior performance in complex combinatorial optimization landscapes.
\paragraph{Differential Evolution (DE)} Operating through vector population mutation and recombination, DE implements a robust evolutionary strategy for continuous optimization domains \cite{DE2019}. This approach generates new candidate solutions by applying weighted differential mutations between population vectors, followed by crossover operations that recombine parental and trial components. Its mutation mechanism provides self-organizing coordinate system rotation and translation invariance, ensuring robust performance across diverse optimization scenarios.
\subsubsection{Large Language Model-based Methods}
\paragraph{FunSearch} A sophisticated optimization methodology leveraging LLMs and a novel island-based evolutionary search paradigm \cite{romera-paredesMathematicalDiscoveriesProgram2024b}. By decomposing the search process into multiple autonomous computational domains exploring solution spaces in parallel, this approach facilitates robust algorithmic discovery. Each computational island generates candidate solutions through strategic generation and evaluation mechanisms, with intermittent knowledge transfer preventing local optimization convergence.

\paragraph{EoH} An advanced computational framework for automated heuristic generation integrating LLMs with evolutionary computation \cite{liuEvolutionHeuristicsEfficient2024a}. The methodology concurrently evolves algorithmic conceptualization and implementation, emulating expert-level heuristic design strategies. Employing a quintuple of sophisticated prompting strategies across exploration and modification taxonomies, it systematically synthesizes diverse and computationally efficient heuristic algorithms.

\paragraph{ReEvo} A pioneering hyper-heuristic optimization framework utilizing LLMs-driven evolutionary search with introspective refinement mechanisms, ReEvo \cite{yeReEvo2024}. By synthesizing evolutionary computation principles with the reflective capabilities of generative models, this approach enables progressive algorithmic enhancement through multi-dimensional strategic and tactical refinement processes.

\subsection{Configuration of MetaHeuristic Baselines}
\label{appendix: baselines}

All meta-heuristic baselines (GA, MA, DE, SA, PSO) are implemented using the \textbf{mealpy (3.0.3)} library to ensure a standardized and reproducible comparison \cite{2023mealpy}. A unified experimental protocol is strictly followed, which includes an identical termination budget (epoch=80, population=45, run=100).

To guarantee the statistical robustness and determinism of our results, we employ a deterministic set of 20 seeds defined \emph{a priori} by an arithmetic progression: \( S = \{ s_0 + \Delta \cdot i \mid i = 0, \dots, 19 \} \) with \( s_0 = 17 \) and \( \Delta = 73 \). This rule-based construction prevents cherry-picking while preserving broad coverage of pseudorandom initializations. All (algorithm, instance) pairs use every seed in \( S \); no seed is added, removed, or replaced after observing any results.

To ensure fairness and mitigate parameter sensitivity, a single set of hyperparameters is fixed for each algorithm and applied across all problems, which avoids iterative per-instance tuning. Key deviations from the library's default parameters, determined through a preliminary calibration, are summarized in Table~\ref{tab:hp-full}.

\begin{table}[h]
  \centering
  \scriptsize
  \caption{Core hyperparameters (Defaults vs.\ Adopted).}

  \label{tab:hp-full}
  \begin{tabular}{l l c c}
    \toprule
    Algorithm & Parameter & Default & Adopted \\
    \midrule
    \multirow{2}{*}{GA}  
         & $p_c$ (crossover rate) & 0.95    & 0.85 \\
         & $p_m$ (mutation rate)  & 0.025   & 0.05 \\
    \midrule
    \multirow{3}{*}{PSO}
         & $c_1$ (cognitive)      & 2.05    & 1.8 \\
         & $c_2$ (social)         & 2.05    & 1.8 \\
         & $w$ (inertia)          & 0.4     & 0.6 \\
    \midrule
    \multirow{3}{*}{DE} 
         & $w_f$ (mutation factor)& 0.1     & 0.6 \\
         & $c_r$ (crossover rate) & 0.9     & 0.9 \\
         & strategy               & 0       & 5 \\
    \midrule
    \multirow{4}{*}{MA}
         & $p_c$(crossover rate)  & 0.85    & 0.85\\
         & $p_m$(mutation rate)   & 0.15    & 0.15  \\
         & $p_{\text{local}}$(local search probability)     & 0.5     & 0.5 \\
         & max\_local\_gens (local search depth)       & 10      & 10 \\
    \midrule
    \multirow{2}{*}{SA} 
         & $T_0$ (initial temperature)  & 100     & 500 \\
         & step\_size (perturbation step size)           & 0.1     & 0.2 \\
    \bottomrule
  \end{tabular}
\end{table}

This standardized setup, with its deterministic seed strategy and fixed hyper-parameters, ensures that performance differences are attributable to the core search strategies rather than implementation or tuning biases.

\subsection{Relaxation Strategy Effectiveness Validation}
The solution space in COPs often exhibits highly fragmented characteristics, with the feasible domain partitioned into isolated connected components. This experiment defines a foundational search algorithm to explore the solution space under different relaxation strategies, aiming to validate the topological structure reconstruction effects of various relaxation approaches. Specifically, we evaluate the performance differences between AutoCO design, expert-defined, and non-relaxation baseline strategies in alleviating solution space fragmentation.

\subsubsection{Base Random Walk Search Algorithm}

To eliminate the inherent search algorithm's influence on solution space exploration, we design a heuristic-free baseline, namely random walk search algorithm (the details are shown in Algorithm \ref{alg:random_walk_exploration}).
\begin{table}[h]
\centering
\caption{Constraint relaxation strategies for SFL8.}
\label{tab:Relaxation strategies8}
\scriptsize
\setlength{\tabcolsep}{2pt}
\begin{tabular}{l*{4}{c}}
\toprule
Strategy & Overlap & Position Range & Safety Area & Global Boundary \\
\midrule
AutoCO-Designed &1.5 & 1.8 &1.2&1.3\\ 
Expert-Designed &2.0 & 1.1 &1.3&2.2\\
No Relaxed  &1.0 & 1.0 &1.0&1.0\\
\bottomrule
\end{tabular}
\vspace{-1em}
\end{table}

\begin{table}[h]
\centering
\caption{Constraint relaxation strategies for SFL5 (Dual).}
\label{tab:Relaxation strategies5}
\scriptsize
\setlength{\tabcolsep}{2pt}
\begin{tabular}{l*{4}{c}}
\toprule
Strategy & Overlap & Position Range & Safety Area & Global Boundary \\
\midrule
AutoCO-Designed & 2.0 & 2.6 & 1.5 & 2.2 \\ 
Expert-Designed & 1.8 & 2.0 & 1.4 & 2.5 \\
No Relaxed & 1.0 & 1.0 & 1.0 & 1.0 \\
\bottomrule
\end{tabular}
\vspace{-0.5em}
\end{table}

\subsubsection{Experimental Configuration}

The experimental configuration is structured to ensure comprehensive and reliable results: 1) We generate a uniformly random pre-generated initial solution set to maintain baseline consistency across all strategies; 2) By using an identical random seed for solution generation, we ensure equivalent and controlled initial conditions for each comparative approach; 3) We consistently evaluate feasibility using original strict constraints, implementing a 30-second search limit per iteration to standardize computational resources; 4) To establish statistical robustness, we conduct 1000 trials for each configuration, systematically exploring computational budgets from 500 to 2500 steps, which enables a rigorous and multifaceted assessment of the proposed relaxation strategies' performance and reliability.

\subsubsection{Relaxation Strategy Configurations}

We evaluate relaxation strategies across two datasets: SFL8 and SFL5 (dual), comparing AutoCO, expert-designed, and safe region relaxation approaches. Tables \ref{tab:Relaxation strategies8} and \ref{tab:Relaxation strategies5} illustrate the detailed relaxation strategies for each dataset.
%\subsubsection{c.4.4. Experimental Results}

\begin{algorithm}[!htp]
\caption{Base Random Walk Search Algorithm}
\label{alg:random_walk_exploration}
\begin{algorithmic}[1]
\Require Initial solution ${S}_0 \in {S}$, Relaxation parameter $\Theta \in {R}^+$, Exploration steps $T \in {N}$
\Ensure Feasible solution discovery count $\eta$, Optimal solution $\mathcal{S}^* \in {S}$, Optimal cost $C^* \in {R}^+$

\Comment{${S}$: Solution Space, $\Theta$: Constraint Relaxation Threshold}

\State Initialize: $\mathcal{S} \gets \mathcal{S}_0$, $\eta \gets 0$, $\mathcal{S}^* \gets \emptyset$, $C^* \gets +\infty$

\For{$t = 1$ \textbf{to} $T$}
    \Comment{Check current solution feasibility under strict constraints}
    \If{$\phi_{\text{strict}}(\mathcal{S})$}
        \State $\eta \gets \eta + 1$
        \State $C \gets \omega(\mathcal{S})$ \Comment{Calculate solution cost}
        \If{$C < C^*$}
            \State $\mathcal{S}^* \gets \mathcal{S}$
            \State $C^* \gets C$
        \EndIf
    \EndIf
    
    \Comment{Generate candidate move via atomic exploration}
    \State $\mathcal{S}' \gets \xi(\mathcal{S})$ \Comment{$\xi$: Atomic move function}
    
    \Comment{Accept move based on relaxed constraints}
    \If{$\phi_{\text{relaxed}}(\mathcal{S}', \Theta)$}
        \State $\mathcal{S} \gets \mathcal{S}'$
    \EndIf
\EndFor

\State \Return{$\eta, \mathcal{S}^*, C^*$}

\Comment{Atomic Move Function $\xi(\mathcal{S})$}
\Function{$\xi$}{$\mathcal{S}$}
    \State $r \gets \text{RandomRectangle}()$ \Comment{Randomly select rectangle}
    \State $\delta \gets \text{RandomDirection}()$ \Comment{Random move direction}
    
    \State $\mathcal{S}_{\text{new}} \gets \text{Copy}(\mathcal{S})$
    \State $\mathcal{S}_{\text{new}}[r].x \gets \mathcal{S}[r].x + \delta.dx$
    \State $\mathcal{S}_{\text{new}}[r].y \gets \mathcal{S}[r].y + \delta.dy$
    
    \State \Return $\mathcal{S}_{\text{new}}$
\EndFunction
\end{algorithmic}
\end{algorithm}

%\begin{table}[h]
%\centering
%\scriptsize  
%\setlength{\tabcolsep}{4pt}
%\caption{The success rates of different strategies under varying computational budgets for SFL8 and SFL5(dual) (The success rate is higher better ↑).}
%\label{tab:strategy-success-rates}
%\begin{tabular}{lcc cccc}
%\toprule
%\textbf{Problem} & \textbf{Strategy} & \textbf{500} & \textbf{1000} & \textbf{1500} & \textbf{2000} & \textbf{2500} \\
%\midrule
%\multirow{3}{*}{SFL8} 
%& No Relaxed & 0.0\% & 22.0\% & 35.0\% & 52.0\% & 52.0\% \\ 
%& AutoCO-Designed & 0.0\% & 35.0\% & 52.0\% & 66.0\% & 80.0\% \\
%& Expert-Designed & 0.0\% & 30.0\% & 40.0\% & 57.0\% & 73.0\% \\
%\hline
%\multirow{3}{*}{SFL5(dual)}
%& No Relaxed & 0.0\% & 12.0\% & 25.0\% & 28.0\% & 28.0\% \\
%& AutoCO-Designed  & 0.0\% & 20.0\% & 45.0\% & 53.0\% & 53.0\% \\
%& Expert-Designed & 0.0\% & 15.0\% & 28.0\% & 42.0\% & 42.0\% \\
%\bottomrule
%\end{tabular}
%\end{table}

\end{document}